\documentclass[lettersize,journal]{IEEEtran}
\usepackage{amsmath,amsfonts}
\usepackage{algorithmic}
\usepackage{array}
\usepackage[caption=false,font=normalsize,labelfont=rm,textfont=rm]{subfig}\usepackage{textcomp}
\usepackage{stfloats}
\usepackage{url}
\usepackage{verbatim}
\usepackage{graphicx}
\def\BibTeX{{\rm B\kern-.05em{\sc i\kern-.025em b}\kern-.08em
    T\kern-.1667em\lower.7ex\hbox{E}\kern-.125emX}}
\usepackage{balance}

\usepackage{xcolor}
\usepackage{algorithm}
\usepackage{times}
\usepackage{soul}
\usepackage{amsthm}
\usepackage{booktabs} % For formal tables
\usepackage{color}

\usepackage{algorithm}
\usepackage{algorithmic}
\usepackage{multirow}
\usepackage{makecell}
\usepackage{mathrsfs}

\usepackage{xspace}
\usepackage{bm}
\usepackage{enumitem}
\usepackage{pifont}% http://ctan.org/pkg/pifont
\usepackage[switch]{lineno}
\usepackage{comment}
\usepackage{mathtools}
\usepackage{cleveref}

\usepackage{longtable}
\usepackage{caption}
\usepackage{graphicx}

\usepackage{wrapfig}
\usepackage{picinpar}
\usepackage{cutwin}
\usepackage[figuresright]{rotating}
\usepackage{cite}

\newcommand{\eg}{\emph{e.g., }}
\newcommand{\ie}{\emph{i.e., }}
\newcommand{\etc}{\emph{etc. }}
\newcommand{\baby}{PDE\textsc{der}\xspace}
\newcommand{\tabincell}[2]{\begin{tabular} {@{}#1@{}}#2 \end{tabular}}

\DeclareSymbolFont{boldoperators}{OT1}{cmr}{bx}{n}
\SetSymbolFont{boldoperators}{bold}{OT1}{cmr}{bx}{n}

\makeatletter
\def\UrlAlphabet{%
      \do\a\do\b\do\c\do\d\do\e\do\f\do\g\do\h\do\i\do\j%
      \do\k\do\l\do\m\do\n\do\o\do\p\do\q\do\r\do\s\do\t%
      \do\u\do\v\do\w\do\x\do\y\do\z\do\A\do\B\do\C\do\D%
      \do\E\do\F\do\G\do\H\do\I\do\J\do\K\do\L\do\M\do\N%
      \do\O\do\P\do\Q\do\R\do\S\do\T\do\U\do\V\do\W\do\X%
      \do\Y\do\Z}
\def\UrlDigits{\do\1\do\2\do\3\do\4\do\5\do\6\do\7\do\8\do\9\do\0}
\g@addto@macro{\UrlBreaks}{\UrlOrds}
\g@addto@macro{\UrlBreaks}{\UrlAlphabet}
\g@addto@macro{\UrlBreaks}{\UrlDigits}
\makeatother

\newcolumntype{a}{>{\columncolor{Gray}}c}

\begin{document}

\title{Generalizing Dynamics Modeling More Easily from Representation Perspective}

\author{
Yiming Wang, Zhengnan Zhang, Genghe Zhang, Jiawen Dan, Changchun Li, Chenlong Hu, 	Chris Nugent, Jun Liu, Ximing Li, Bo Yang
\thanks{This work was supported by the National Science and Technology Major Project (No.2021ZD0112500), the National Natural Science Foundation of China (No.62276113). \textit{(Corresponding author: Ximing Li and Bo Yang.)} }

\thanks{Yiming Wang, Zhengnan Zhang, Genghe Zhang, Changchun Li, and Bo Yang are with the College of Computer Science and Technology, and Key Laboratory of Symbolic Computation and Knowledge Engineering of Ministry of Education, Jilin University, China; Jiawen Dan and Chenglong Hu are with the College of Software and Key Laboratory of Symbolic Computation and Knowledge Engineering of Ministry of Education, Jilin University, China; 
Chris Nugent and Jun Liu are with the School of Computing, Belfast, Northern Ireland, UK;
Ximing Li is with the College of Computer Science, and Technology, Key Laboratory of Symbolic Computation and Knowledge Engineering of Ministry of Education, Jilin University, China.
% , and Center for Advanced Intelligence Project, RIKEN, Japan. 
(e-mail: yimingw17@gmail.com, zhengnan0801@gmail.com, genghezhang99@gmail.com, danjiawen19@gmail.com, changchunli93@gmail.com, huchenglong239@gmail.com, cd.nugent@ulster.ac.uk, j.liu@ulster.ac.uk, liximing86@gmail.com,  ybo@jlu.edu.cn).}
}


\markboth{Journal of \LaTeX\ Class Files,~Vol.~18, No.~9, September~2020}%
{How to Use the IEEEtran \LaTeX \ Templates}

\IEEEpubid{0000--0000/00\$00.00~\copyright~2021 IEEE}
% % column for its text to clear the IEEEpubid mark.

\maketitle

\label{1}
\begin{abstract}

Learning system dynamics from observations is a critical problem in many applications over various real-world complex systems, \eg climate, ecology, and fluid systems. Recently, neural dynamics modeling method have become a prevalent solution that embeds the object's observations into a latent space before learning dynamics using neural methods such as neural Ordinary Differential Equations (ODE). Existing dynamics modeling methods induce a specific model for each observation of different complex systems, resulting in poor generalization across systems. Inspired by the great success of pre-trained models, we conduct a generalized \underline{P}re-trained \underline{D}ynamics \underline{E}nco\underline{DER} (\textbf{\baby}) which can embed the original state observations into a latent space where the dynamics can be captured more easily.
To conduct the generalized \baby, we pre-train any Pre-trained Language Model (PLM) by minimizing the Lyapunov exponent objective, which constrains the chaotic behavior of governing dynamics learned in the latent space. By penalizing the divergence of embedded observations, our \baby promotes locally stable and well-structured latent dynamics, thereby facilitating more effective dynamics modeling than in the original observation space. In addition, we incorporate reconstruction and forecasting objectives to mitigate the risk of obtaining an over-smoothed latent space.
Specifically, we collect 152 sets of real-world and synthetic observations from 23 complex systems as pre-training corpora and employ them to pre-train \baby. Given any future dynamic observation, we can fine-tune \baby with any specific dynamics modeling method. We evaluate \baby on 12 dynamic systems by short/long-term forecasting under both in-domain and cross-domain settings, and the empirical results indicate the effectiveness and generalizability of \baby.

\end{abstract}

\begin{IEEEkeywords}
Time series forecasting, dynamics modeling, pre-trained language model, Lyapunov exponent.
\end{IEEEkeywords}

\section{Introduction}
\label{2}

% 1. background

% \begin{figure}
% \centering
% \includegraphics[width=0.5\textwidth]{flowchart.jpg}
% \caption{pic description} \label{fig1}
% \end{figure}

% \begin{wrapfigure}{r}{0.4\textwidth}%靠文字内容的右侧
% % \begin{wrapfigure}{r}{4cm}%靠文字内容的右侧
% \centering
% \includegraphics[width=0.4\textwidth]{flowchart.jpg}
% % \includegraphics[width=4cm]{flowchart.jpg}
% \caption{FLowchart of our \baby.}
% \end{wrapfigure}

\IEEEPARstart{S}{ystem} dynamics, which describes the object states evolving over time, is a powerful methodology to conceptualize complex systems from various domains, \eg climate, ecology, and fluid systems \cite{Biology2006, Microbial2016, Complex2016, Neuronal2014, Fates2019, Collective2018, Nettide2016, PowerSocial2018, Uncovering2019, Dyorigins2019}. In parallel with physical methods, learning system dynamics from abundant observations has drawn much attention, and the neural-based dynamics modeling method, such as neural Ordinary Differential Equations (ODE) become the representative \cite{NODE2018}.  

To our knowledge, the basic idea of the representative method is to embed the original states of objects into a latent space before learning the dynamics using neural-based methods and finally followed by a decoder \cite{NDCN2020}. Although the existing methods have been successfully applied in various systems, most of them must train a specific model given observations from different systems, resulting in limited generalizability. To meet this challenge, several recent studies investigate generic methods that can simultaneously handle multiple dynamics from various systems and environments \cite{yin2021leads,wang2022meta,kirchmeyer2022generalizing,GGODE2023}. These methods enhance the model generalization mostly in a multi-task learning manner, where they decompose the dynamics modeling module into two components, one is shared across environments and the other is an environment-specific module. Nevertheless, these methods mainly concentrate on systems sharing the same parameterized equation forms with different parameters of initial conditions, \eg Heat diffusion systems with different diffusion rates. 
Unfortunately, the governing equations in real-world systems are almost unknown, the strategy may limit the generalization capacity when facing unknown systems. Besides, the potentially huge gap between various dynamics challenges the development of generic dynamics modeling methods, making it an open problem.

% % These strategies may limit the generalization capacity when facing unknown parameters and systems with different parameterized equations. 
% Unfortunately, the governing equations in real-world systems are almost unknown and the potentially huge gap between various dynamics challenges the development of generic dynamics modeling methods, making it an open problem.
% % Unfortunately, due to the potential huge gap between various dynamics, developing generic dynamics modeling methods faces significant challenges and is still an open problem.

\IEEEpubidadjcol

Inspired by the great success of pre-trained models, we raise a question: {Instead of developing generic dynamics modeling methods, whether we can conduct a generalized \textbf{P}re-trained \textbf{D}ynamics \textbf{E}nco\textbf{DER} (\textbf{\baby}), which, for various complex systems, can embed their original states into a latent space, where the dynamics can be captured more easily.}
To this end, we propose the basic idea that if there exist a latent space where the governing dynamics learned from the embedded observations exhibit less chaos, such dynamics could be captured more readily than in the original space. We therefore resort to the Lyapunov exponent to measure the chaos degree of approximated dynamics by quantifying the exponential divergence of observation embeddings \cite{liapounoff1907probleme}. 
Specifically, we pre-train \baby by minimizing the Lyapunov exponent objective on the embedded observations to constrain the chaotic behavior of governing dynamics learned in the latent space. 
This encourages the encoder to induce representations that can capture more stable and well-structured dynamics, thereby enabling more effective and tractable latent dynamics modeling than in the original observation space,  meanwhile preserving essential dynamic behaviors. In addition, we incorporate reconstruction and forecasting objectives as auxiliary tasks to enhance forecasting performance and meanwhile preventing the latent space from becoming over-smoothed, which could harm the representation capacity for dynamics modeling.

We achieve \baby using Pre-trained Language Models (PLM) and can employ any PLM to further update it over dynamics observations by tokenization techniques. We collect 152 sets of real-world and synthetic observations from 23 complex systems as pre-training corpora.
With the pre-trained \baby, we can fine-tune \baby with any specific dynamics modeling method on any future dynamics observations. We present an example of fine-tuning \baby by a black-box GNN-based dynamics leaner to learn specific dynamics and evaluate \baby by fine-tuning on 12 different dynamics systems under both in-domain and cross-domain forecasting tasks. The results indicate the effectiveness and generalizability of \baby when learning specific dynamics.

The major contributions of our paper are listed below:
\begin{itemize}
    \item We propose to pre-train a generalized dynamics encoder by minimizing the Lyapunov exponent to induce a stable latent space where governing dynamics could be captured more easily and effectively than in the original space. We also incorporate reconstruction and forecasting objectives as auxiliary tasks to prevent an over-smoothed latent space, enabling the optimized \baby to be readily adopted for downstream dynamics modeling when directly modeling in the original space is intractable.
    % In this manner, the optimized \baby encoder could be adopted for further dynamics modeling when modeling in the original observation space is intractable.

    % By minimizing the Lyapunov exponent, the complex systems learnt in the latent space are less chaotic. Therefore, the optimized encoder could be adopted for further learning specific dynamics when learning dynamics in the original observation space is intractable. We also set reconstruction loss and forecasting loss as auxiliary objectives to enhance the model forecasting capability.
    \item We collect extensive complex-system observations from diverse domains as pre-training corpora, comprising 152 sets from 13 synthetic and 10 real-world dynamics systems. The diversity of pre-training corpora is ensured by drawing from multiple domains and varying parameters when generating observations from synthetic system.
    
    \item We illustrate an example of fine-tuning \baby to learn specific dynamics using a GNN-based dynamics learner and conduct several empirical studies to evaluate dynamics modeling performance by short/long-term forecasting under both in-domain and cross-domain settings.

\end{itemize}

% In a nutshell, the major contributions of this paper can be outlined below:

% \begin{itemize}
%     \item We propose a novel idea of updating PLM to build a generic encoder \baby by minimizing Lyapunov exponent, which can make dynamics modeling more easily.
%     % \item We propose a novel idea of updating PLM by forecasting, reconstruction and minimizing the Lyapunov exponent to build a generic encoder \baby for dynamics modeling.

%    % \item We introduce the minimization of maximum Lyapunov exponent in the latent space which facilitates modeling dynamics with less chaos.

%     \item We collect extensive real-world and synthetic observations from various complex systems to train \baby.

%     \item We conduct numbers of experiments to evaluate \baby on short/long-term forecasting under both in-domain and cross-domain settings. 
% \end{itemize}

The rest of the paper is organized as follows: Section \ref{2} reviews related work. Preliminaries on Ordinary Differential Equations and the Lyapunov exponent are presented in Section \ref{3}. We describe our methodology in Section \ref{4}, including \baby pre-training and fine-tuning for learning specific dynamics. Experimental settings and results are reported in Section \ref{5}. Conclusions and limitations are given in Section \ref{6}. Detailed calculations of the maximal Lyapunov exponent and non-invariance proof under \baby encoding-decoding paradigm are provided in Appendices \ref{App_mle} and \ref{App_proof}.

\section{Related Work}
\label{2}

In this section, we introduce the relevant literature about neural dynamics modeling methods and time series forecasting methods using pre-trained language models.

% \subsection{Dynamics Models}
\paragraph{Dynamics Modeling Methods}

Equipped with the parameterized neural ODE \cite{NODE2018}, recent dynamics modeling methods primarily fall into the data-driven manner. \cite{NDCN2020} is a pioneer combining the neural ODE with GNN \cite{wu2020comprehensive} to approximate continuous-time dynamics of networks at an arbitrary time on the interaction graph. 
\cite{GNS2020} approximates dynamics by modeling temporal graphs and object states in the latent space simultaneously.
Furthermore, to model the irregularly sampled partial observations, LGODE \cite{LG-ODE2020} combines the temporal self-attention mechanism to model the irregular timeline by positional encoding in the transformer architecture. 
% Similarly, \cite{liu2022seeing} proposes to incorporate transformer architecture into modeling dynamics with interacting objects. 
Based on LGODE, several methods have been proposed to handle the scenarios of dynamic graphs, cross-domain environments, introducing manual interventions and incorporating physics knowledge by time-reversal symmetry regularization term \cite{CG-ODE2021,GGODE2023,CAG-ODE2024,huang2023tango}. 
\cite{CG-ODE2021} models copuled dynamics of co-evolving nodes and edges with GNN-based ODEs in the continuous-environment dynamics by describing dynamics of different dynamics through shared ODEs and environment-specific factors.  
\cite{huang2023tango} enhanced modeling capability of physical systems through time-reversal symmetry regularization term, improving the forecasting accuracy and robustness for complex systems.
\cite{CAG-ODE2024} adopts GNN-based ODE function to capture the dynamic effects of treatments over time and the combined effects of multiple treatments. 
\cite{PGODE2024} proposed a novel graph ODE model that significantly enhances modeling capability and generalization performance of multi-agent dynamics through the introduction of contextual prototypes.

Apart from GG-ODE\cite{GGODE2023}, there also exist several generalizable dynamics modeling methods.
% To induce generalizable dynamics modeling methods,
\cite{wang2021physics} incorporates prior physics knowledge into existing deep learning approaches to enhance generalizability on unseen environments. \cite{kirchmeyer2022generalizing} handles generalizable dynamics modeling by adjusting model on domain-specific context parameters. Nevertheless, most existing methods enhance generalizability by resorting to multi-task paradigm, where they assume the dynamics modeling module consists of a cross-environment sharing module and an environment-specific module. They mainly focus on systems sharing same parameterized equation forms with different parameters or initial conditions. These assumptions limit generalizability on unknown systems and cross-domain scenarios. 

\textit{Distinctions}: Existing methods can only generalize on environment-level, where the parameterized equation forms of dynamics are available. While our \baby can generalize from the representation level, making it feasible to generalize across both environment and system on any unknown dynamics.

\paragraph{Pre-trained Language Models for Time Series Forecasting}
Recently, PLMs, especially transformer-based PLMs, have achieved promising performance in NLP \cite{PLM4NLPsurvey} and have been successfully transferred into time series forecasting task domain due to the sequence modeling capacity of transformers \cite{PM4TSFsurvey}.
% Recently, PLMs have been successfully applied to time series forecasting task due to the sequence modeling capacity of transformers.
\cite{Patching2023} proposes to segment the time series into patches and model the sequence dependencies by transformers.
\cite{gruver2024large} encodes time series as a string of numbers, forecasting the next token in the text to achieve sequence forecasting results, allowing time series input to adapt unilaterally to the input format of language models. 
% PromptCast \cite{xue2023promptcast} converts the numerical inputs and outputs of time series into prompts, constructing forecasting tasks in a sentence-by-sentence manner. 
AutoTimes \cite{liu2024autotimes} transforms time series data into a format understandable by LLMs for auto-regressive forecasting. LLM4TS \cite{chang2024llm4ts} proposes a two-phase fine-tuning method, first aligning the model with the characteristics of the time series to better adapt the LLM, and then further fine-tuning the model guided by downstream forecasting tasks in the second phase. Time-LLM \cite{TimeLLM2024} requires no fine-tuning of any layers in the LLM, simply freezing the LLM and using two learnable modules to process inputs and outputs. Additionally, \cite{zhou2023one} provides a unified framework for various time series tasks. \cite{UP2ME2024} proposed a framework that transitions from univariate pre-training to multivariate fine-tuning. 
% Through self-supervised pre-training and cross-channel dependency fine-tuning, it demonstrates excellent performance across various time series tasks.
\cite{niu2025langtime} proposed temporal comprehension prompts to integrate semantics and channel details for multi-modal time series forecasting. 
\cite{zhongtime2025} adopts pre-trained Vision-Language Models (VLMs) to extract temporal, visual and contextual information from three aspects.

To our knowledge, handling dynamics modeling by PLMs have not been fully explored. Therefore, we adopt PLMs to embed dynamics observations into generalizable representations to utilize the sequence modeling capacity for inducing the latent space where dynamics could be captured more easily.

% By resorting to PLMs, our \baby can learn generalizable representations and approximate dynamics more easily and efficiently benefiting from the chaotic behavior constraints.
% forecasting capacity and pre-training paradigm.

% leverage the forecasting capacity of transformer and incorporates the generalizability from the pre-training and fine-tuning paradigm.

% the pre-training fine-tuning paradigm 

% . A
% Our \baby resorts to PLM 

% By resorting to forecast time series by PLMs, we 

% \cite{NODE2018,CG-ODE2021}

% \cite{VideoLLaMA2023}

% \cite{SpeechGPT2023}]

% \cite{Blip2023,Flamingo2022,MiniGPT42024}

% \cite{zhou2023one,gruver2024large,TimeLLM2024,liu2024autotimes}

% [\cite{xue2023promptcast, Patching2023, liu2024autotimes, TimeLLM2024}]
% PatchTST[\cite{zhou2023one}]
% [\cite{zhang2022self}]
% [\cite{UP2ME2024}]
\section{Preliminaries}
\label{3}

In this section, we introduce the notations of dynamics modeling, Ordinary Differential Equations (ODE) for dynamics modeling and Lyapunov exponent.

\begin{figure*}[!t]
% \begin{figure}[H]
    \centering
    \includegraphics[width=0.9\linewidth]{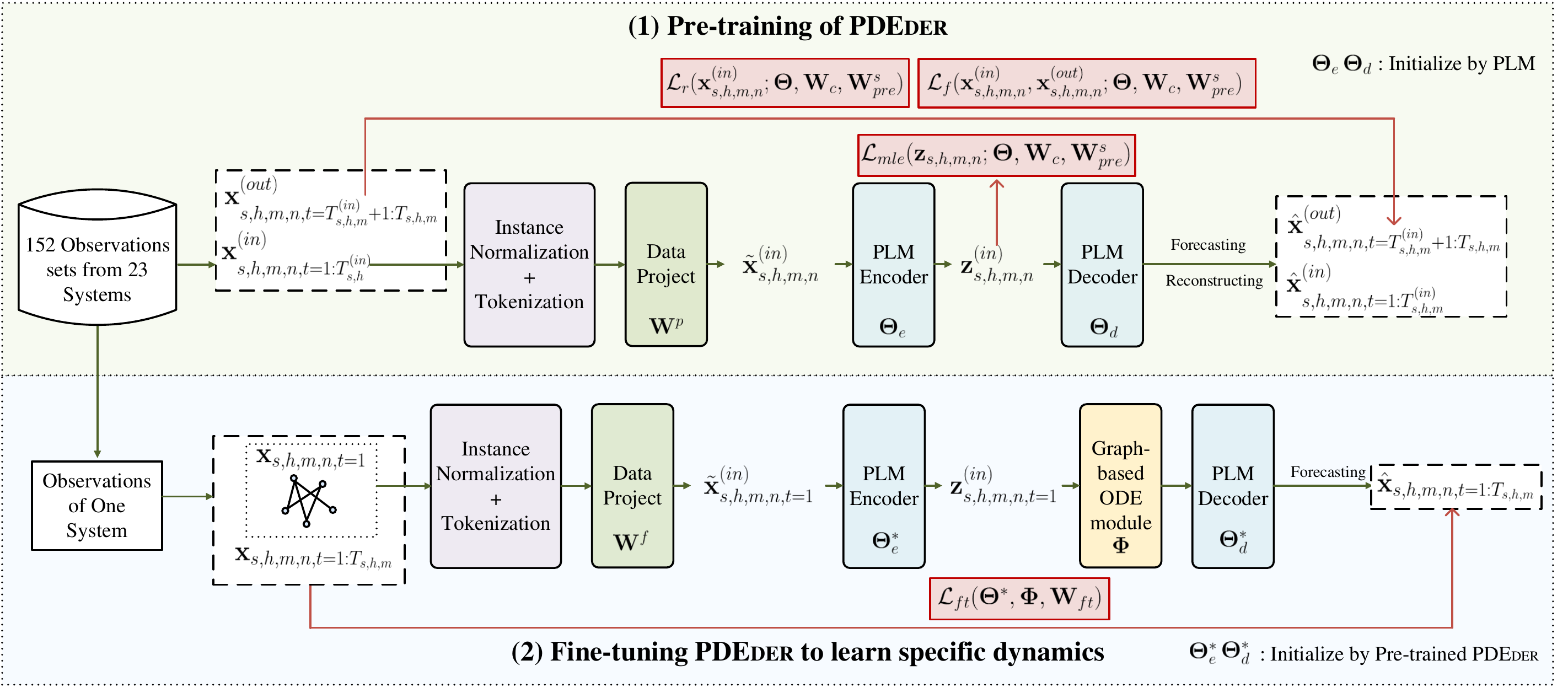} 
    \caption{The overall pipeline of pre-training (top) and fine-tuning (bottom) \baby. (1) Pre-training of \baby on large-scale dynamics observations: The sequences are first tokenized and normalized, followed by projection with system-specific data projection layers to align the dimension. The projected sequences are then encoded into latent space and decoded for reconstruction and forecasting. To pre-train \baby, we employ a maximal Lyapunov exponent objective on sequence embeddings along with reconstruction and forecasting objectives.    
    (2) Fine-tuning \baby to approximate specific dynamics: The first token of an observation sequence is taken as the initial state and encoded into the latent space via the pre-trained encoder of \baby. Then we approximate governing dynamics by integrating using the embedded initial states within the latent space through a GNN-based ODE module. Finally, we decode the integrated solutions by the pre-trained decoder and fine-tuning \baby by minimizing the forecasting objective against the ground-truth observation sequence.
    % (2) Fine-tuning \baby to approximate dynamics on a specific system: We adopt the first token of an observation sequence as the initial state and encode it into the latent space using the pre-trained encoder of \baby. The governing dynamics are then approximated by integrating on the embedded initial states within the latent space through a GNN-based ODE module. Finally, the integrated solutions are decoded by the pre-trained decoder and the entire model is fine-tuned by minimizing the forecasting against the ground-truth observation sequence.
    }
    \label{fg_overall}
\end{figure*}
% \end{figure}

% version 3.0 2025.1.17
Commonly speaking, a dynamics consists of a set of interacting objects whose states co-evolve over an interacting graph along timeline. It could be formalized into a graph $\mathcal{G}=(\mathcal{V},\mathcal{E})$, where $\mathcal{V}=\{\mathbf{x}_n\}_{n=1}^N$ is the set of $N$ interacting objects and $\mathcal{E}=\{\mathbf{e}_{i,j}\}_{i,j=1}^N$ denotes the interactions among them. The observations of each object $\mathbf{x}_n$ is a trajectory of states along time $T$. On time $t$, the state of object $n$ can be represented as a vector $\mathbf{x}_{n,t}\in\mathbb{R}^{V}$ where $V$ is the system-specific state dimension. In general, the evolution of object states are governed by some hidden regularities. Given the states observations $\mathcal{G}$, we aim to extract the hidden governing dynamics and can forecast the states at an arbitrary time $t$.

% For object $n$, $\mathbf{x}_n=\{\mathbf{x}_{n,t}\}_{t=1}^{T}$ denotes the trajectory of observations and $\mathbf{x}_{n,t}\in\mathbb{R}^{V}$ denotes the observed state on time $t$; $V$ is the system-specific dimension. When modeling the hidden dynamics, we are given the states observations of each object and the interaction graph among them 

% $\mathcal{V}=\{\mathbf{x}_n\}_{n=1}^N$, the states of each object co-evolve along with timeline. Along $T$ timestamps, we can obtain the observed trajectories for each object $\mathbf{x}_n=\{\mathbf{x}_{n,t}\}_{t=1}^{T}$, where $\mathbf{x}_{n,t}\in\mathbb{R}^{V}$ denotes the observed state at time $t$; $V$ is the system-specific state dimension. $\mathcal{E}=\{\mathbf{e}_{i,j}\}_{i,j=1}^N$ denotes the interactions among them. For 

% Consider a dynamical system $\mathcal{G}=(\mathcal{V},\mathcal{E})$ with $N$ interacting objects $\mathcal{V}=\{\mathbf{x}_n\}_{n=1}^N$ and $\mathcal{E}=\{\mathbf{e}_{i,j}\}_{i,j=1}^N$ denotes the interactions among them. $\mathbf{x}_n=\{\mathbf{x}_{n,t}\}_{t=1}^{T}$ denotes the observation trajectories of each object and $\mathbf{x}_{n,t}\in\mathbb{R}^{V}$ denotes the observed feature on time $t$. Our goal is to learn a model $f$ which can model the hidden dynamics law from historical observations and predict future states for any object.  

\paragraph{Ordinary Differential Equations (ODE) for learning governing dynamics}

Conventionally, the evolution of each object states in a dynamics can be described by Ordinary Differential Equations (ODE): $\dot{\mathbf{x}}_{n,t} \coloneqq \frac{d\mathbf{x}_{n,t}}{dt}=g(\mathbf{x}_{1,t}, \ldots, \mathbf{x}_{N,t};\mathcal{G})$, where $g(\cdot;\mathcal{G})$ is a hand-crafted function to model the characteristic from observations. The differential equations describe the instantaneous state changing rates of each object under mutual influences. Given the initial states of each object $\{\mathbf{x}_{1,t=1}, \ldots, \mathbf{x}_{N,t=1}\}$, state at an arbitrary time $\tau$ can be calculated by integrating the differential equation over timeline:
\begin{equation}
    \mathbf{x}_{n,\tau} = \mathbf{x}_{n,1} + \int_1^\tau g(\mathbf{x}_{1,t}, \ldots, \mathbf{x}_{N,t};\mathcal{G})dt.
\end{equation}
The above integration is also called an ODE initial value problem \cite{Initial2021} for this differential equation, which could be solved by numerical ODE solvers such as Euler's method, Dormand-Prince DOPRI5 \cite{DOPRI52018}, Runge-Kutta \cite{RungeKutta2019}, \etc The dynamics can be approximated with these numerical methods at an arbitrary time.

% Ordinary differential equations (ODE) for multi-agent dynamic system

% We first tokenize and normalize the observation sequences. Then we adopt a system-specific data projection layer to align different sequence dimensions. The projected sequence are then encoded into latent representations and decoded for reconstruction and forecasting. We adopt a maximal Lyapunov Exponent objective on the embedded observations to pre-train \baby. The reconstruction and forecasting objectives are also adopted as auxiliary. 

% (2) Fine-tuning the pre-trained \baby to learn specific dynamics using GNN-based ODE module on a set of observations with different initial values under identical system. Specially, the GNN-based ODE module could be substituted by any dynamics learner.

% We first use the first token of an observation sequence as initial state and encode by the pre-trained \baby into latent space. Then we approximate the governing dynamics by integrating in the latent space using a GNN-based ODE module. Finally, we decode the solutions by the pre-trained decoder and fine-tune with minimizing the forecasting loss against ground-truth observation sequence.

\paragraph{Lyapunov Exponent of Complex Systems}

Lyapunov exponent measures the chaotic degree of a complex system by quantifying the exponential divergence of observations with close initial values \cite{eckmann1985ergodic,farmer1987predicting,sano1985measurement,wolf1985determining}. Two trajectories in the phase space with initial separation vector $\bm\delta_0$ diverge at a rate given by:
\begin{equation}
    |\bm\delta(t)| \approx e^{\lambda t}  |\bm\delta_0|,
\end{equation}
where $\lambda$ is the Lyapunov exponent. According to the definition, when $\lambda>0$, the dynamics is chaotic and sensitive to initial values; when $\lambda<0$, the system is periodic and trajectories with nearby initial states will finally consolidate. Generally, we apply the maximal Lyapunov exponent (\textbf{M-Lyapunov exponent}) to measure thedegree of chaoshaos in a complex system, defined as follows: 
% Lyapunov characteristic exponent is a widely applied metric which can estimate the chaos in the dynamical system by quantifying the exponential divergence of dynamical observations with close initial values \cite{eckmann1985ergodic,farmer1987predicting,sano1985measurement,wolf1985determining}. 
% Given the observations of a one-dimensional dynamics system $\mathbf{x}_{t+1} = F(\mathbf{x}_{t})$, the Lyapunov exponent is defined as:
% Given a dynamics $\mathbf{x}_{t+1} =f(\mathbf{x})$, MLE is defined as:
\begin{equation}
    % \lambda = \lim_{t\rightarrow\infty} \frac{1}{t}\sum_{i=0}^{t-1}
    % % \ln\left| \frac{df(\mathbf{x})}{d\mathbf{x}} \right|_{\mathbf{x}=\mathbf{x}_i}.
    % \ln \left|f^\prime(\mathbf{x}_i)\right| .
    \lambda = \lim_{t\rightarrow\infty} \lim_{|\bm\delta_0|\rightarrow0} \ln \frac{|\bm\delta(t)|}{|\bm\delta_0|}.
\end{equation}
% According to the definition, when $\lambda>0$, the dynamical system is chaotic and sensitive to initial values. The trajectories will diverge with nearby initial states; when $\lambda<0$, the system is periodic and trajectories with nearby initial states will finally consolidate.
% % ; when $\lambda=0$, the system is marginally stable. 
% % The most commonly used methods for calculating Lyapunov exponent include Wolf method \cite{wolf1985determining}, Jacobian method \cite{eckmann1985ergodic,eckmann1986liapunov}, p-norm method \cite{barna1993new} and small-dataset method \cite{rosenstein1993practical}.
% % The small-dataset method \cite{rosenstein1993practical} is widely used for calculating the maximum Lyapunov exponent due to the reliability for small datasets and implementation efficiency and simplicity. This method quantifying the exponent by the divergence over observation nearest neighbor pairs along timeline. 
Given a set of observations $\mathbf{x}$ from a complex system, we can employ the \textit{small-dataset method} to estimate its M-Lyapunov exponent, defined as $\lambda = \ell_{mle}(\mathbf{h})$. Detailed calculation is presented in Appendix \ref{App_mle}. 

\section{Methodology}\label{4}

% \subsection{Overview of \baby}
% \subsection{Overview of the Pre-trained Dynamic Encoder (\baby)}
\subsection{Overview of the Pre-trained Dynamic Encoder}

\baby consists of an encoder and decoder to induce a latent space for dynamics modeling.
In this section, we introduce the pre-training of \baby and fine-tuning \baby to learn specific dynamics. The overall pipeline is presented in Figure \ref{fg_overall}.

% In pre-training, we use massive collected dynamics observations to pre-train the encoder and decoder. In fine-tuning to learn specific dynamics, we encode 
(1) \textbf{Pre-training}: To achieve generalizability, we first collect massive observations from diverse synthetic and real-world systems as pre-training corpus. We encode the collected observations by the encoder, and decode to reconstruct and forecast. Then we pre-train \baby by minimizing the M-Lyapunov exponent on the embedded observations. We also introduce reconstruction and forecasting objectives on the decoded observations to enhance the model forecasting capacity.
% Then we pre-train \baby with these collected observations by minimizing the M-Lyapunov Exponent on the embeddings of each observations. We also introduce reconstruction and forecasting objectives to enhance the model forecasting capacity.
% Then we pre-train \baby with these collected observations on three tasks: i) \textit{reconstructing observed sequence}, ii) \textit{forecasting future states}, iii) \textit{minimizing the maximum Lyapunov exponent of observed sequence}. 
By introducing the Lyapunov objective, we can guide \baby to induce embeddings which can capture more stable and well-structured dynamics learned in the latent space, 
% can learn dynamics with less chaos, 
thereby facilitating the formation of latent space where the dynamics characteristics could be captured more easily. 
% \ie minimizing the Lyapunov exponent of observed sequence, reconstructing observed sequence and forecasting future states. The intuition is that if \baby can By reducing the chaos of dynamics system learnt from the embeddings in the latent space, \textbf{The basic assumption is if \baby can accurately forecast various systems and induce embeddings which can learn systems with less chaos, it can form a latent space where the dynamics can be captured more easily.}
% \textbf{A basic assumption is that if \baby can accurately forecast various systems, it forms a latent space where the dynamics can be captured more easily. And dynamics systems with less chaos are more easily captured in the embedding space}. 

(2) \textbf{Fine-tuning}: With the pre-trained \baby, we can generate dynamics-enriched embeddings for initial states and approximate dynamics by fine-tuning \baby on each specific system. 
The pre-trained encoder serves as the representation learner to embed initial states into latent space. And then we can adopt any general dynamics learner to approximate dynamics by solving the initial value problem in the latent space and finally decode the solutions back to the original space by the pre-trained decoder. We fine-tune \baby by minimizing the forecasting loss between the decoded solutions and ground-truth sequence.
We give an example of fine-tuning \baby by a GNN-based neural method.

\begin{table*}[!ht]
    \centering
    % \small
    % \scriptsize
    % \footnotesize
    \tiny
    \caption{Statistics of collected dynamics. $N_{s,h,m}$ is the number of objects; $T_{s,h,m}$ is the length of observation sequence; $V_s$ is the feature dimension; $M_{s,h}$ is the number of generated samples; $H_s$ is the number of different parameter settings.}    
    % set numbers of observations correspond with different hyper-parameters.}
    \renewcommand\arraystretch{1.0}
    % \begin{tabular}{p{65pt}<{\centering}| p{40pt}<{\centering} p{80pt}<{\centering} p{80pt}<{\centering} p{30pt}<{\centering} p{30pt}<{\centering}}
    % \resizebox{\linewidth}{!}{    \begin{tabular}{p{90pt}<{\centering}| p{40pt}<{\centering}p{35pt}<{\centering} p{90pt}<{\centering} p{65pt}<{\centering} p{25pt}<{\centering} p{25pt}<{\centering}p{25pt}<{\centering}}
    \resizebox{\linewidth}{!}{
    \begin{tabular}{c| ccccccc}
    % \Xhline{1.2pt}
    \toprule
    System      &Type   &Domain       &$N_{s,h,m}$      &$T_{s,h,m}$   &$H_s$  &$M_{s,h}$      &$V_s$  \\
    % \hline
    \midrule
	Charged \cite{NRI2018}            &Synthetic	&Physics	&$\{$5,10,15,20$\}$	&$\{$400,600$\}$	&8	&5000	&4	\\
	Springs \cite{NRI2018}           &Synthetic	&Physics	&$\{$20,25,30,35,40$\}$	&$\{$200,300$\}$	&10	&3500	&4	\\
	Mutualistic Interaction Dynamics \cite{Complex2016}        &Synthetic	&Physics	&$\{$100,121,169,196,225$\}$	&$\{$300,350,400$\}$	&15	&1500	&1	\\
	Heat Diffusion \cite{Heat2012}              &Synthetic	&Physics	&$\{$225,256,289,324$\}$	&$\{$200,250,300$\}$	&12	&1500	&1	\\
	1D Diffusion-Reaction \cite{PDEBench2022}  &Synthetic	&Fluid	&$\{$256,368,464,512$\}$	&$\{$200,225,250,275$\}$	&16	&700	&1	\\
	% \makecell[c]{1D Compressible\\Navier-Stokes} 
    % 1D-CFD 
    1D Compressible Navier-Stokes
    \cite{klaasen1984stationary} &Synthetic	&Fluid	&$\{$300,350,400$\}$	&$\{$600,625$\}$	&6	&300	&3	\\
	% \makecell[c]{2D Compressible\\Navier-Stokes}   
    % 2D-CFD 
    2D Compressible Navier-Stokes
    \cite{klaasen1984stationary} &Synthetic	&Fluid	&$\{$400,625,784,1024$\}$	&$\{$100,150,200$\}$	&12	&200	&4	\\
	Burgers \cite{PDEBench2022}           &Synthetic	&Fluid	&$\{$400,425,450$\}$	&$\{$512,768,960,1024$\}$	&12	&250	&1	\\
	Advection \cite{PDEBench2022}          &Synthetic	&Fluid	&$\{$500,550$\}$	&$\{$700,725,750$\}$	&6	&500	&1	\\
	Darcy Flow \cite{PDEBench2022}         &Synthetic	&Fluid	&$\{$625,676$\}$	&$\{$400,425,450$\}$	&6	&700	&1	\\
	Gene Regulatory \cite{Complex2016}              &Synthetic	&Biology	&$\{$729,841,900,1024$\}$	&$\{$125,150,175,200$\}$	&16	&500	&1	\\
	% Shallow-Water \cite{PDEBench2022}     &Synthetic	&Fluid	&768	&500	&1	&3000	&1	\\
	2D Diffusion-Reaction \cite{PDEBench2022}  &Synthetic	&Fluid	&900	&120	&1	&5000	&2	\\
	Diffusion-Sorption \cite{PDEBench2022}     &Synthetic	&Fluid	&1024	&101	&1	&10000	&1	\\
									
	LA \cite{choi2023climate}                &Real-world	&Climate	&274	&384	&-	&1	&10	\\
	SD \cite{choi2023climate}                &Real-world	&Climate	&282	&384	&-	&1	&10	\\
	NYCTaxi \cite{ATFM2021}           &Real-world	&Traffic	&75	&17520	&-	&1	&2	\\
	CHIBike \cite{wang2021libcity}           &Real-world	&Traffic	&270	&4416	&-	&1	&2	\\
	TDrive \cite{ATFM2021}            &Real-world	&Traffic	&1024	&3600	&-	&1	&2	\\
	PEMS03 \cite{chen2001freeway}            &Real-world	&Traffic	&358	&26208	&-	&1	&1	\\
	PEMS04 \cite{chen2001freeway}            &Real-world	&Traffic	&307	&16992	&-	&1	&3	\\
	PEMS07 \cite{chen2001freeway}            &Real-world	&Traffic	&883	&28224	&-	&1	&1	\\
	PEMS08 \cite{chen2001freeway}            &Real-world	&Traffic	&170	&17856	&-	&1	&3	\\

    NOAA \cite{hwang2021climate}               &Real-world  &Climate 
    &\tabincell{c}{$\{$17,24,27,29,40,40,40,46,49,53,55,65,\\77,89,93,108,160,179,199,216,225,253$\}$}
    % [17,24,27,29,40,40,40,46,49,53,55,65,77,89,93,108,160,179,199,216,225,253]         
    &7305     &-      &22     &3\\ 
    \bottomrule
    % \Xhline{1.5pt}
    \end{tabular}}
    \label{T_datasets}
\end{table*}

% \subsection{Pre-training of the Pre-trained Dynamic Encoder (\baby)}
\subsection{Pre-training of \baby}

In this section, we first introduce the collection of pre-training corpora, and each step of pre-training pipeline.

\paragraph{Pre-training Corpora Collection}

We first introduce the collection of dynamics observations. We collect 152 sets of observations as pre-training corpora, including generating 121 synthetic sets from 13 systems with various parameters, and collecting 31 sets of real-world observations from 10 systems. Their domains consist of physics, fluid, biology, climate and traffic. When generating observations for synthetic systems, we set different parameters (\eg system-specific parameters, object number and sequence lengths) for each system and generate multiple samples for each parameter set by assigning different random initial values. Here, each sample consists of observation sequences of several (non-)interacting objects. Specially, we varied the number of samples to balance the total observation amounts for each system. Statistics of collected dynamics observations are presented in Table \ref{T_datasets}.

% (\ie each with observations of (non-)interacting objects governed by identical dynamics) for each parameter set by assigning different random initial values. Statistics of collected dynamics observations are presented in Table \ref{T_datasets}.

\textit{Example}: For system ``Charged'', we set 4 numbers of objects $\{5,10,15,20\}$ and 2 sequence lengths $\{400,600\}$. For each combination of these two parameters, we generate 5000 samples with different initializations. Then we have 8 parameters settings, each with 5000 sets of observations. The two parameters are varied from $[5, 1024]$ for all systems. 
% The observation statistics and detailed introductions of all systems are presented in Appendix \ref{App_data}.

\textit{Notations}: For each of $S$ synthetic systems $s\in[S]$, we set $H_s$ different parameter settings, including numbers of objects and sequence lengths. For each parameter setting $h$, we generate $M_{s,h}$ samples with different initial values $\{\mathcal{G}_{s,h,m}=(\mathcal{V}_{s,h,m},\mathcal{E}_{s,h,m})\}_{m=1}^{M_{s,h}}$, where $\mathcal{V}_{s,h,m} = \{\mathbf{x}_{s,h,m,n}\}_{n=1}^{N_{s,h,m}}$ denotes the observations of $N_{s,h,m}$ objects and $\mathcal{E}_{s,h,m}=\{<\mathbf{x}_{s,h,m,i}, \mathbf{x}_{s,h,m,j}>\}_{i,j\in[N_{s,h,m}]}$ denotes the interactions among them. The observation $\mathbf{x}_{s,h,m,n}\in\mathbb{R}^{T_{s,h,m}\times V_s}$ of object $n$ is denoted as a states trajectory along time $T_{s,h,m}$, where $V_s$ denotes the system-specific state dimension. 
We split $\mathbf{x}_{s,h,m,n}$ into two sub-observations $\mathbf{x}_{s,h,m,n}^{(in)} = \{\mathbf{x}_{s,h,m,n,t}\}_{t=1}^{t=T_{s,h,m}^{(in)}}$ and $\mathbf{x}_{s,h,m,n}^{(out)} = \{\mathbf{x}_{s,h,m,n,t}\}_{t=T_{s,h,m}^{(in)}+1}^{t=T_{s,h,m}}$ to pre-train \baby on multiple tasks. 
By ingesting $\mathbf{x}_{s,h,m,n}^{(in)}$, we pre-train by minimizing the M-Lyapunov exponent on the embeddings of $\mathbf{x}_{s,h,m,n}^{(in)}$, reconstructing $\mathbf{x}_{s,h,m,n}^{(in)}$ and forecasting $\mathbf{x}_{s,h,m,n}^{(out)}$. 

\textit{Index omission}: For brevity, in the rest of the paper, we omit the subscripts of the system index $s$ and sample index $h$ for some variables, \eg we use $\mathbf{x}_{m,n}$ to represent $\mathbf{x}_{s,h,m,n}$.

% and use all combinations of these two parameters to generate 8 sets of observations. 

% We generate 8 sets of observations using all combinations of the two sets of parameters. Each of the 8 sets corresponds with different system-specific hyper-parameters. Then we generate 5000 observation sequences under each parameter setting with random initial values.
% For all systems, we vary the number of objects and sequence length both from $[5, 1024]$.
% % and try our best to {\color{red}cover each 100 interval}. 
% The statistics of observations and systems are presented in Appendix \ref{App_data} and Table \ref{T_datasets}.
% % The statistics of observations are illustrated in Table \ref{T_datasets} and the detailed descriptions of each system are presented in Appendix \ref{App_data}. 
% To pre-train \baby on multiple tasks, We split the observations $\mathbf{x}_{m,n}$ into two sub-observations $\mathbf{x}_{m,n}^{(in)} = \{\mathbf{x}_{m,n,t}\}_{t=1}^{t=T_k}$ and $\mathbf{x}_{m,n}^{(out)} = \{\mathbf{x}_{m,n,t}\}_{t=T_{k}+1}^{t=T_m}$. By ingesting the $\mathbf{x}_{m,n}^{(in)}$, we learn \baby by reconstructing $\mathbf{x}_{m,n}^{(in)}$ and forecasting $\mathbf{x}_{m,n}^{(out)}$. 

% \subsubsection{Sequence Tokenization}
\paragraph{Tokenization}
We then pre-train \baby with the collected observations.
To adapt the input observations with various lengths and serve as input tokens for transformer-based PLMs, we tokenize the input sequence following \cite{Patching2023}. 
% version3.0 2025.5.12
% To reconstruct the phase space for calculating the Lyapunov exponent and adapt the input observations to transformer-based PLMs, we tokenize the input sequence following \cite{Patching2023} and adopt the tokens as time delay embeddings.
For object $n$, we patch the input observation ${\mathbf{x}}_{m,n}^{(in)}\in\mathbb{R}^{T_{m}^{(in)}\times V_s}$ into 
% $\overline{\mathbf{x}}_{m,n}^{(in)} \in \mathbb{R}^{L_p \times P_m}$, 
$\overline{\mathbf{x}}_{m,n}^{(in)} \in \mathbb{R}^{P_{m} \times L_p \times V_s}$, where $L_p$ denotes the patch length; $P_{m} = \lfloor \frac{(T_{m}^{(in)} - L_p)}{R} \rfloor + 2$  denotes the number of patches; $R$ denotes the stride. In this manner, the trajectory lengths are reduced by $R$ times, which can simultaneously maintain the local semantics in long-term dynamics modeling and significantly reduce the space and time costs in training.
Besides, to benefit domain adaptation and generalization, we also introduce Instance Normalization to handle the distribution shifts among various domains following \cite{InsNorm2021}.

\paragraph{Data Projection}
To handle dimension diversity of states across different systems, we adopt a system-specific data projection module to align observations by mapping into the dimension of PLM. For the patched tokens $\overline{\mathbf{x}}_{m,n}^{(in)}$, we first flatten it into  $\overline{\mathbf{x}}_{m,n}^{(in)(fl)} \in \mathbb{R}^{P_{m} \times (L_p \cdot V_s) }$, and then project it by a linear layer $\tilde{\mathbf{x}}_{m,n}^{(in)} = f_{\mathbf{W}_{dp}^{s}}(\overline{\mathbf{x}}_{m,n}^{(in)(fl)})$, where $\mathbf{W}_{dp}^{s}\in\mathbb{R}^{D \times (L_p \cdot V_s)}$ denotes the system-specific trainable parameters; $D$ denotes the hidden dimension of PLM.

% \subsubsection{Pre-trained LLM}
% \subsubsection{Encode Patched Sequence with Pre-trained LLM}
% \paragraph{Learn with PLM}
\paragraph{Encoder \& Decoder}

For object $n$, we first encode the tokenized and projected sequence $\tilde{\mathbf{x}}_{,m,n}^{(in)}$ into embeddings $\mathbf{z}_{m,n}$ by a convolutional layer $f_{\mathbf{W}_c}(\cdot)$
% following \cite{LLM4TS2023}, 
and the encoder of a PLM $f_{\bm\Theta_e}(\cdot)$. Then, we decode $\mathbf{z}_{m,n}$ by the decoder of a PLM $f_{\bm\Theta_d}(\cdot)$ attached by two flatten-linear layers $f_{\mathbf{W}_{r}^{s}}(\cdot)$ and $f_{\mathbf{W}_{f}^{s}}(\cdot)$, which serves for reconstruction and forecasting, respectively. Detailed calculations are as listed below:
\begin{align}
% \begin{equation}
% \begin{split}    
    \mathbf{z}_{m,n} &= f_{\bm\Theta_e}(f_{\mathbf{W}_c}(\tilde{\mathbf{x}}_{m,n}^{(in)})), \label{Eq_enc} \\
    \hat{\mathbf{x}}_{m,n}^{(in)} &= f_{\mathbf{W}_r^{s}}(f_{\bm\Theta_d}(\mathbf{z}_{m,n})), \label{Eq_dec1} \\
    \hat{\mathbf{x}}_{m,n}^{(out)} &= f_{\mathbf{W}_f^{s}}(f_{\bm\Theta_d}(\mathbf{z}_{m,n})). 
    % \label{Eq_dec2}
    % \end{split}
% \end{equation}
\end{align}
\paragraph{Pre-training Objective}
% version 3.0
To induce a latent space which can learn dynamics more easily, we pre-train \baby by minimizing the M-Lyapunov exponent of observation embeddings, along with reconstruction and forecasting losses. Given the embeddings $\mathbf{z}_{m,n}$,
% of each input observation  $\mathbf{x}_{m,n}^{(in)}$, 
the objective is given as:
\begin{equation}
    \mathcal{L}_{mle}(\mathbf{z}_{m,n}; \bm\Theta,\mathbf{W}_c,\mathbf{W}_{pre}^s) = \ell_{mle}(\mathbf{z}_{m,n}),
\end{equation}
where $\ell_{mle}(\cdot)$ denotes the exponent estimated by small-dataset method \cite{rosenstein1993practical} and the detailed calculation is presented in Appendix \ref{App_mle};
% the small-dataset method to estimate M-Lyapunov exponent proposed \cite{rosenstein1993practical} and the detailed calculation is presented in Appendix \ref{App_mle};
% \footnote{We apply an off-the-shelf tool of the small-dataset method, which is available at \url{https://github.com/manu-mannattil/nolitsa}.}
$\bm\Theta=\{\bm\Theta_e, \bm\Theta_d\}$ denotes the parameters set of the encoder/decoder of a PLM; $\mathbf{W}_{pre}^s = \{\mathbf{W}_{dp}^{s}, \mathbf{W}_r^{s}, \mathbf{W}_p^{s}\}$ are the parameters of system $s$ corresponds with $\mathbf{x}_{m,n}$. 
Additionally, we introduce reconstruction loss against the input sequence $\mathbf{x}_{m,n}^{(in)}$ and forecasting loss against the future sequence $\mathbf{x}_{m,n}^{(out)}$ to prevent over-smoothed latent space and meanwhile enhance forecasting capacities:
\begin{equation}
\begin{split}    
    \mathcal{L}_{r}(\mathbf{x}_{m,n}^{(in)};\bm\Theta, \mathbf{W}_c, \mathbf{W}_{pre}^s) &= 
    \sum_{t=1}^{T_{m}^{(in)}}\ell_1(\hat{\mathbf{x}}_{m,n,t}^{(in)}, \mathbf{x}_{m,n,t}^{(in)}),\\
    \mathcal{L}_{f}(\mathbf{x}_{m,n}^{(in)},\mathbf{x}_{m,n}^{(out)};\bm\Theta, \mathbf{W}_c,\mathbf{W}_{pre}^s) &= 
    \!\! \sum_{t=T_{m}^{(in)}+1}^{T_{m}} \!\ell_1(\hat{\mathbf{x}}_{m,n,t}^{(out)}, \mathbf{x}_{m,n,t}^{(out)}),
\end{split}
\end{equation}
where $\ell_1(\cdot)$ denotes the L1 loss which treats all errors equally \cite{TSLoss2024}. Therefore, the overall pre-training objective for each $\mathbf{x}_{m,n}$ is defined as:
\begin{equation}
\begin{split}    
    &\mathcal{L}_{pre}(\mathbf{x}_{m,n};\bm\Theta, \mathbf{W}_c,\mathbf{W}_{pre}) = \\ &\quad\sum_{n=1}^{N_{m}}
    \Big(\mathcal{L}_{mle}(\mathbf{z}_{m,n};\bm\Theta, \mathbf{W}_c,\mathbf{W}_{pre}^s) \\
    &\quad\quad + \rho_1 \cdot \mathcal{L}_{r}(\mathbf{x}_{m,n}^{(in)};\bm\Theta,\mathbf{W}_c, \mathbf{W}_{pre}^s)\\
    &\quad\quad + \rho_2 \cdot\mathcal{L}_{f}(\mathbf{x}_{m,n}^{(in)},\mathbf{x}_{m,n}^{(out)};\bm\Theta,\mathbf{W}_c, \mathbf{W}_{pre}^s)\Big),
\label{Eq_obj1}
\end{split}
\end{equation}
% \begin{equation}
% \begin{split}    
%      \mathcal{L}_p(\bm\Theta, \mathbf{W}_c,\mathbf{W}_{pre}) = \\
%     \sum_{s=1}^{S} \sum_{z=1}^{H_s} \sum_{m=1}^{M_{s,h}} \sum_{n=1}^{N_{s,h,m}} \;
%     &\left(\mathcal{L}_{mle}(\mathbf{h}_{s,h,m,n};\bm\Theta, \mathbf{W}_c,\mathbf{W}_{pre}^s) + \rho_1 \cdot \mathcal{L}_{rec}(\mathbf{x}_{s,h,m,n}^{(in)};\bm\Theta,\mathbf{W}_c, \mathbf{W}_{pre}^s) \right.\\
%     &\ \ \left.+ \rho_2 \cdot\mathcal{L}_{fore}(\mathbf{x}_{s,h,m,n}^{(in)},\mathbf{x}_{s,h,m,n}^{(out)};\bm\Theta,\mathbf{W}_c, \mathbf{W}_{pre}^s)\right),
% \label{Eq_obj1}
% \end{split}
% \end{equation}
where $\mathbf{W}_{pre} = \{\mathbf{W}_{dp}^{s}, \mathbf{W}_r^{s}, \mathbf{W}_p^{s}\}_{s=1}^S$; $\{\rho_1,\rho_2\}$ denote the scaling parameters. We pre-train \baby using all the observations from all collected systems.

Besides, the maximal Lyapunov exponent invariance under invertible transformations or non-invertible transformations which can map the observation trajectories of two different systems to each other \cite{MLEinvariance2001} is not available under our scenario and we present the proof in Appendix \ref{App_proof}.

\subsection{Approximating Specific Dynamics with \baby}
After pre-training, we can adopt \baby as an embedder and approximate specific dynamics in the latent space by any dynamics learner. In this section, we introduce an example of learning dynamics by fine-tuning \baby with a black-box GNN-based dynamics learner \cite{NDCN2020}.

\begin{algorithm}[!t]
  \footnotesize
  % \caption{Pre-training \baby to learn dynamics-enriched embeddings.}
  \caption{Pre-training procedure of \baby.}
  \label{AlgModel1}
    \begin{algorithmic}[1]
        \REQUIRE{153 Observations sets $\{\mathcal{V}_{s,h,m}\}_{m=1}^{M_{s,h}}$ from $S$ systems} 
        \ENSURE{Optimal parameters of \baby $\{\bm\Theta^*,{\mathbf{W}_{pre}^*}\}$}
        \STATE Initialize $\bm\Theta$ by pre-trained LM;
        \FOR{round $r=1$ to $MaxRound$}
            \STATE Sample 10 systems;
            \FOR{epoch $e=1$ to $MaxEpoch_p$}
                \FOR{iter $it=1$ to $MaxIter_p$}
                    \STATE Sample $B$ observations from 5 systems as a batch;
                    \STATE Encode and decode observations by Eqs.\eqref{Eq_enc}, \eqref{Eq_dec1};
                    % , \eqref{Eq_dec2};
                    \STATE Calculate the pre-training objective of Eq.\eqref{Eq_obj1};
                    \STATE Update $\{\bm\Theta, \mathbf{W}_c,\mathbf{W}_{pre}\}$ by back-propagation;
                \ENDFOR
            \ENDFOR
        \ENDFOR
    \end{algorithmic}
\end{algorithm}

% version2.0 2025.1.18
As mentioned in preliminaries, we can model hidden dynamics of a specific system by solving the ODE initial value problem with initial observations $\{\mathbf{x}_{m,1,1}, \ldots, \mathbf{x}_{m,N_d,1}\}$ of $N_d$ objects. To fit \baby, we adapt the first projected token $\tilde{\mathbf{x}}_{m,n,1}\in\mathbb{R}^{D}$ as the initial value for each object to learn dynamics. We first encode $\tilde{\mathbf{x}}_{m,n,1}$ into latent space by the pre-trained encoder $\mathbf{z}_{m,n,1} = f_{\bm\Theta_e^*}(f_{\mathbf{W}_c}(\tilde{\mathbf{x}}_{m,n,1}))$, where $\bm\Theta_e^*$ denotes the pre-trained parameters of \baby encoder. 
Then we approximate the governing dynamics by solving the initial value problem, \ie integrating over timeline using the embedded initial states $\mathbf{z}_{m,n,1}$ in the latent space. Then we decode the integrated solutions by the pre-trained decoder and fine-tune \baby by minimizing the forecasting loss against real observations.

% Then we can approximate the governing dynamics in the latent space by any specific dynamics learner and decode for re-projection.

% \paragraph{Example: Learn with Black-box Dynamics Learner.}
% \paragraph{Example 1: Fine-tune \baby by a Black-Box Dynamics Learner.}
% \paragraph{Example: Fine-tune \baby by a Black-Box Dynamics Learner.}

% version 1.0
\begin{comment}
To model the dynamics where the objects affect each other along with evolution,
% Since the objects in a multi-agent dynamics system affect each other along with evolution, 
following \cite{NDCN2020}, we adapt a GNN-based module $g(\cdot)$ to model dynamics by incorporating the interactions among objects in the latent space. Let $\mathbf{A}_m \in \mathbb{R}^{N_m \times N_m}$ denotes the adjacent matrix of the interaction graph $\mathcal{G}_m$ and $\mathbf{h}_{m,\cdot ,\tau}=[\mathbf{h}_{m,1,\tau}, \ldots, \mathbf{h}_{m,N_m,\tau}] \in \mathbb{R}^{N_m \times H}$ denotes the embeddings at an arbitrary time $\tau$ $(1<\tau\leq T_m)$, we describe the dynamics by the following ODE:
% the ordinary differential equation at time $t$ is formulated as: 
\end{comment}

% \textbf{Example of Fine-tuning \baby by a Black-Box Dynamics Learner}:
% version2.0 2025.1.18
To handle co-evolving objects, we adapt a GNN-based dynamics leaner to approximate dynamics on interacting graph in the latent space \cite{NDCN2020}. Let $\mathbf{A}_{m} \in \mathbb{R}^{N_{m} \times N_{m}}$ denotes the adjacent matrix of the interacting graph $\mathcal{G}_m$ and $\mathbf{z}_{m,\cdot ,\tau}=[\mathbf{z}_{m,1,\tau}, \ldots, \mathbf{z}_{m,N_m,\tau}] \in \mathbb{R}^{N_{m} \times D}$ denotes the representations encoded by pre-trained \baby at an arbitrary time $\tau$ $(1<\tau\leq T_{m})$. We can describe the governing dynamics by the following ODE:
% the ordinary differential equation at time $t$ is formulated as: 
\begin{equation}
    % \frac{d\mathbf{h}^{m,t}}{dt} = g(\mathbf{h}^{m,t}) = \psi(\mathbf{W}_g^\top\Phi(\mathbf{A_d})\mathbf{h}^{m,t} ),
    \!\dot{\mathbf{z}}_{m,\cdot,\tau} =  \frac{d\mathbf{z}_{m,\cdot ,\tau}}{dt} \!= g_1(\mathbf{z}_{m,\cdot,\tau}) \!=\! \sigma(\mathbf{W}_g^\top \bm\Lambda_{m} \mathbf{z}_{m,\cdot,\tau} ),
\label{Eq_ft_dot}
\end{equation}
where $\bm\Lambda_{m}=\mathbf{D}_{m}^{-\frac{1}{2}}(\mathbf{D}_{m}-\mathbf{A}_{m})\mathbf{D}_{m}^{-\frac{1}{2}}\in\mathbb{R}^{N_{m}\times N_{m}}$ denotes the Laplacian normalization of $\mathbf{A}_{m}$; $\mathbf{D}_{m}$ denotes the degree matrix of $\mathbf{A}_{m}$; $\mathbf{W}_g$ denotes the trainable parameters shared across timeline; $\sigma(\cdot)$ denotes the $\mathrm{ReLU}$ activation function. With the above ODE and initial values $\mathbf{z}_{m,\cdot,1}$, we can model dynamics by integrating over continuous time:
% At an arbitrary time $T$, the corresponding hidden representation can be learnt by integrating $g(\cdot)$ from time $t=0$ to $t=T$:
\begin{equation}
    % \mathbf{h}^{m,T} = \int_{0}^T \psi(\mathbf{W}_g^\top\Phi(\mathbf{A_d})\mathbf{h}^{m,t} ) dt.
    \mathbf{z}_{m,\cdot,\tau} = \mathbf{z}_{m,\cdot,1} + \int_{1}^\tau \sigma(\mathbf{W}_g^\top \bm\Lambda_{m} \mathbf{z}_{m,\cdot,t} ) dt.
\label{Eq_ft_int}
\end{equation}
The hidden representations $\mathbf{z}_{m,\cdot,t}$ for all time points $t\in(1,T_{m}]$ could be calculated with the above integration. Then we reconstruct the states by the pre-trained \baby decoder $\hat{\mathbf{x}}_{m,n} = f_{\mathbf{W}_r^s}(f_{\bm\Theta_d^*}(\mathbf{z}_{m,n}))$ and learn hidden dynamics of system $s$ by minimizing the forecasting loss of all sample $m\in[M_s]$ against the ground-truth observations $\mathbf{x}_{m,n}$: 
% After calculating $\{\mathbf{h}^{m,t}\}_{t=1}^{T_m}$ for time $[1,T_m]$, we can decode the representations by the pre-trained encoder of the foundation model:
% \begin{equation}
%     \hat{\mathbf{x}}^m = f(f(\mathbf{h}^{m}; \mathbf{W}_d);\mathbf{W}_r).
% \end{equation}
% The dynamics model is learnt by minimizing the reconstruction loss against the ground-truth observations $\mathbf{x}^m$. The objective is formulated as below:
\begin{equation}
    % \mathcal{L}_f(\mathbf{W}^f) = \sum_{d=1}^{D} \sum_{m=1}^{M_d} \sum_{n=1}^{N_d} \ell_1(\hat{\mathbf{x}}_n^m, \mathbf{x}_n^m),
    \mathcal{L}_{ft}(\bm\Theta^*, \bm\Phi, \mathbf{W}_{ft}) =  \sum_{m=1}^{M_s} \sum_{n=1}^{N_m} \ell_1(\hat{\mathbf{x}}_{m,n}, \mathbf{x}_{m,n}),
\label{Eq_obj2}
\end{equation}
where $\bm\Theta^*=\{\bm\Theta_e^*, \bm\Theta_d^*\}$ denotes parameters of pre-trained \baby encoder (decoder); $\bm\Phi=\{\mathbf{W}_g\}$ denotes parameters of the neural ODE module; $\mathbf{W}_{ft}=\{\mathbf{W}_c, \{\mathbf{W}_{dp}^s, \mathbf{W}_r^s\}_{s=1}^S\}$.

\begin{algorithm}[!t]
  \footnotesize
  % \caption{Fine-tuning \baby to learn specific dynamics.}
  % \caption{Fine-tuning of \baby with GNN-based dynamics learner.}
  \caption{Fine-tuning \baby to learn specific dynamics.}
  \label{AlgModel2}
    \begin{algorithmic}[1]
        \REQUIRE{Observations $\{\mathcal{G}_{s,h,m}=(\mathcal{V}_{s,h,m}, \mathcal{E}_{s,h,m})\}_{m=1}^{M_{s,h}}$ of system $s$, pre-trained \baby encoder and decoder $\bm\Theta^*$ } 
        % \ENSURE{Optimal parameters of approximated dynamics $\mathbf{W}_g$}
        \ENSURE{Optimal parameters of approximated dynamics $\bm\Phi$}
        % \ENSURE{\color{red}Analytical solution or optimal parameters $\mathbf{W}_g$ of approximated dynamics} 
        \STATE Initialize $\bm\Theta$ by $\bm\Theta^*$; 
        \FOR{epoch $e=1$ to $MaxEpoch_f$}
            \FOR{iter $it=1$ to $MaxIter_f$}
                \STATE Encode initial values by the pre-trained encoder Eq.\eqref{Eq_enc};
                \STATE Calculate fine-tuning objective of Eq.\eqref{Eq_obj2};
                \STATE Solve the initial value problem by Eq.\eqref{Eq_ft_int};
                \STATE Decode integrated values by the pre-trained decoder Eq.\eqref{Eq_dec1};
                % \STATE Calculate integration for each time point by Eq.\eqref{Eq_ft_int} {\color{red} Or solve the initial value problem?};
                \STATE Calculate the fine-tuning objective of Eq.\eqref{Eq_obj2};
                \STATE Update $\{\mathbf{\bm\Theta^* }, \bm\Phi, \mathbf{W}_{ft}\}$ by back-propagation;
            \ENDFOR
        \ENDFOR
    % \RETURN approximated dynamics.
    \end{algorithmic}
\end{algorithm}

\subsection{Model Training Details} We first pre-train \baby on all collected dynamics observations (without graph) by Eq.\eqref{Eq_obj1} for $E_p$ epochs. To balance the massive observations from different systems, we randomly sample $5$ systems for each training round and form batches by the ratios of samples amounts. As for fine-tuning \baby to learn specific dynamics, we adopt a set of observations for one system and fine-tune \baby with Eq.\eqref{Eq_obj2} for $E_f$ epochs. The training details are presented in Algorithms \ref{AlgModel1} and \ref{AlgModel2}.
% The details of pre-training and fine-tuning \baby are presented in Algorithm \ref{AlgModel1} and \ref{AlgModel2}.
% in Appendix \ref{App_alg}.

% To handle the massive observations and various numbers of samples on different systems, we randomly choose $10$ systems for each training round and pre-train \baby for $5$ epochs with all the observations from these systems. When learning a specific dynamics, we fine-tune \baby with Eq.6 for $E_f$. The details of pre-training and fine-tuning \baby are presented in Algorithm \ref{AlgModel1} and \ref{AlgModel2}.

\section{Experiments}
\label{5}

\subsection{Experimental Settings}
\paragraph{Datasets}
% [need to introduce the ranges in the settings of number of nodes and length of sequence. try to cover all circumstances]
% [sort the datasets in the increasing order of number of nodes]

% In fine-tuning, we adopt 17 dynamics owning object interactions for validation, including 7 sets of synthetic observations: Mutualistic, Heat Diffusion, 2D Compressible Navier-Stokes, Darcy Flow, Gene, Shallow Water, 2D Diffusion Reaction; and 10 real-world observations: LA, SD, TDrive, CHIBike, NYCTaxi, PEMS03, PEMS04, PEMS07, PEMS08 and NOAA. Detailed descriptions are introduced in Appendix \ref{App_data}.
In fine-tuning, we adopt 12 dynamics with interacting objects for validation, including 4 sets of synthetic observations: Mutualistic, Heat Diffusion, 2D Compressible Navier-Stokes (2D-CFD) and Gene; 8 set of real-world observations: T-Drive, CHIBike, NYCTaxi, PEMS03, PEMS04, PEMS07, PEMS08 and NOAA. 
% Detailed descriptions are introduced in Appendix \ref{App_data}.

% We adopt 153 datasets including 122 simulated datasets generated from 14 dynamics systems with various hyper-parameters and 31 realistic datasets from 10 domains for examination. To cover all possible circumstances, we randomly set the number of nodes $N_d$ and lengths of sequences $T_m$ from the range of $[100,1000]$. The statistics of datasets are illustrated in Table \ref{T_datasets} and the detailed descriptions are presented in Appendix \ref{App_data}. 

\paragraph{Baselines}
% NDCN 2020\cite{NDCN2020}
% LG-ODE 2020 \cite{LG-ODE2020}
% CG-ODE 2021	\cite{CG-ODE2021}
% ST-GODE 2021 \cite{STGODE2021}	
% TG-ODE 2024	\cite{TGODE2024}
% MT-ODE 2022

We apply 6 baseline methods for comparison, including 1 single-agent dynamics modeling method LatentODE \cite{LatentODE2021}; 3 multi-agent methods GNS\cite{GNS2020}, NDCN \cite{NDCN2020}, TREAT \cite{huang2023tango}; and 2 multivariate time series forecasting methods ST-GODE \cite{STGODE2021}, MT-GODE \cite{MTGODE2022}. 
Specially, due to the model setting of STGODE and MTGODE whose input and output lengths should be identical for training and testing, we only compare with them on real-world systems.
Details of baseline methods are introduced below:
% in Appendix \ref{App_baseline}. 
% Following \baby, we adopt instance normalization on observations for all methods. Details of baseline methods are listed in Appendix \ref{App_baseline}:
% We apply 6 baseline methods for comparison, including 1 single-agent dynamics modeling method LatentODE \cite{LatentODE2021}; 3 multi-agent methods GNS\cite{GNS2020}, NDCN \cite{NDCN2020}, TANGO \cite{huang2023tango}; and 2 multivariate time series forecasting methods ST-GODE \cite{STGODE2021}, MT-GODE \cite{MTGODE2022}. Details of these methods are listed below:
\begin{itemize}
\item{\textbf{LatentODE}}\footnote{\url{https://github.com/wanjiaZhao1203/TREAT}\label{TREAT}} \cite{LatentODE2021} is a single-agent dynamics modeling method without considering any object interactions. 
% We adopt the default parameter settings.
% is a dynamics modeling method modeling the observation sequences independently without considering any interacting graphs. We adopt the default provided hyper-parameters.
\item{\textbf{GNS}}\footnote{\url{https://github.com/zhouxian/GNS-PyTorch}} \cite{GNS2020} is a discrete GNN-based dynamics modeling method handling temporal graphs. We modified the graph learning module into static graph structures.
\item{\textbf{NDCN}}\footnote{\url{https://github.com/calvin-zcx/ndcn}} \cite{NDCN2020} combines the ODEs with GNNs and approximate the integration differential equation systems by the GNN module. We adopt the same initial states setting as our \baby, which are introduced in Section \ref{para_setting}.
% The last state in training sequence is adopted as the initial states for testing.
% In testing, we set the state of the last time point of training observations as the initial states to forecast the test observations. We set the state of the last time point of training observations as the initial states for testing.
\item{\textbf{STGODE}}\footnote{\url{https://github.com/square-coder/STGODE}} \cite{STGODE2021} incorporates the geometry spatial information into continuous dynamics learning for multivariate time series forecasting.
% STGODE constructs two types of graphs, including spatial and semantic correlations to capture the spatial temporal semantics by a continuous GNN with residual connections.
\item{\textbf{MTGODE}}\footnote{\url{https://github.com/TrustAGI-Lab/MTGODE}} \cite{MTGODE2022} solves the multivariate time series forecasting by mapping the interacting observations into dynamic-graph and solve by learning continuous spatial-temporal dynamics in latent space. 
% We adopt the multi-step forecasting setting.
\item{\textbf{TREAT}}\textsuperscript{\ref{TREAT}} \cite{huang2023tango} introduce time-reversal symmetry into GNN-based ODE learner and models the observations and reversed observations simultaneously. We adopt identical graphs for all time points.
% For all time points, we use the same static graph the same as our \baby.
\end{itemize}

\paragraph{Implementation and Forecasting Settings} 
\label{para_setting}
We pre-trained two versions of \baby and conducted examinations on both versions. Specifically, the corresponding PLM modules are initialized by pre-trained {\ttfamily{T5}} and {\ttfamily{T5-small}}, denoted as ``\baby'' and ``\baby-small'', respectively. We implement our \baby by PyTorch and run it on 2 GeForce RTX 4090 GPUs in a Ubuntu platform with 128G memories.

% We adopt the pre-trained {\ttfamily{T5}} to initialize the PLM module ``\baby'' and ``\baby-small'' denotes the versions using pre-trained {\ttfamily{T5}} and {\ttfamily{T5-small}}. 
% We implement our \baby by PyTorch
% % \footnote{\url{https://pytorch.org}} 
% and run it on 2 GeForce RTX 4090 GPUs in a Ubuntu platform with 128G memories.

\textit{Pre-training Settings}: We apply all available 152 sets of observations for pre-training. To handle the massive observations, we randomly sample $5$ system sets for each training round and form batches by the sample number proportion of each system. We train $E_p=5$ epochs for each round. The learning rates are set as $1e-3$ for the PLM module and $1e-2$ for rest parameters. The patch length and stride are set as $25$ and $6$, respectively. To align observations with different lengths, we split each observation by a look-back window of length $150$. The input sequence length $T_\cdot^{(in)}$is set as $100$ for all systems, \ie ingest the former sequence of length 100 and decode to reconstruct the former sequence and forecast the latter sequence of length 50. 
% The scaling parameter $\rho$ in the objective is set as $1$.

% \begin{sidewaystable*}[!t]
\begin{table*}[!t]
    \centering
    % \scriptsize
    \setlength{\tabcolsep}{0.1pt}
    \tiny	
    \caption{Full results of dynamics forecasting varying forecasting lengths. Each score is the averaged results of multiple sample with different initial values. The best scores are in \textbf{boldface} and the second best scores are in \underline{underline}. ``Forecast Length'' denotes the forecasting length. For synthetic systems, we truncate the test sequence by ratios of $\{10\%,20\%,50\%,70\%,80\%,100\%\}$ to form short/long-term forecasting. ``Avg'' denotes the averaged results of all forecasting lengths. MAPE is in \%. ``-'' means that results are not available.  ``\baby'' and ``\baby-small'' denotes the versions using pre-trained {\ttfamily{T5}} and {\ttfamily{T5-small}}. (Due to the model setting of STGODE and MTGODE that the input and output lengths should be identical for training and testing, we only compare with them on realistic systems.)}
    \renewcommand\arraystretch{1}
    \resizebox{\linewidth}{!}{
    \begin{tabular}{p{18pt}<{\centering}p{30pt}<{\centering}|  p{20pt}<{\centering}|
    p{18pt}<{\centering} p{18pt}<{\centering} p{18pt}<{\centering} |
    p{18pt}<{\centering} p{18pt}<{\centering} p{18pt}<{\centering} |
    p{18pt}<{\centering} p{18pt}<{\centering} p{18pt}<{\centering} |
    p{18pt}<{\centering} p{18pt}<{\centering} p{18pt}<{\centering} |
    p{18pt}<{\centering} p{18pt}<{\centering} p{18pt}<{\centering} |
    p{18pt}<{\centering} p{18pt}<{\centering} p{18pt}<{\centering} ||
    p{18pt}<{\centering} p{18pt}<{\centering} p{18pt}<{\centering} |
    p{18pt}<{\centering} p{18pt}<{\centering} p{18pt}<{\centering}  }    
    \toprule

    \multicolumn{3}{c|}{Method Type} 
    &\multicolumn{3}{c|}{Single-Agent Method}    
    &\multicolumn{9}{c|}{Multi-Agent Methods}    
    &\multicolumn{6}{c||}{Multivariate Time Series Forecasting Methods}    
    &\multicolumn{6}{c}{Our Method}    \\

    % \cmidrule{4-27}
    % \toprule
    \midrule
    
    % \multirow{2}{*}{System} 
    \multicolumn{2}{c|}{\multirow{2}{*}{System}} 
    &\multirow{2}{*}{\makecell{Forecast\\Length}} 
    &\multicolumn{3}{c|}{LatentODE \cite{LatentODE2021}}    
    &\multicolumn{3}{c|}{GNS \cite{GNS2020}}    
    &\multicolumn{3}{c|}{NDCN \cite{NDCN2020}}
    % &\multicolumn{3}{c|}{LGODE}    
    &\multicolumn{3}{c|}{TREAT \cite{huang2023tango}}
    &\multicolumn{3}{c|}{STGODE \cite{STGODE2021}}    
    &\multicolumn{3}{c||}{MTGODE \cite{MTGODE2022}}    
    &\multicolumn{3}{c|}{\baby} 
    &\multicolumn{3}{c}{\baby-small} 
    \\
    & &
    &RMSE &MAE &MAPE
    &RMSE &MAE &MAPE
    &RMSE &MAE &MAPE
    &RMSE &MAE &MAPE
    &RMSE &MAE &MAPE
    &RMSE &MAE &MAPE
    &RMSE &MAE &MAPE
    &RMSE &MAE &MAPE \\

    \midrule
    \multirow{28}{*}{\rotatebox{90}{Synthetic Systems}}
    &\multirow{7}{*}{Mutualistic}
    &10\% 	
    &2.706 	&2.541 	&32.56		
    &1.383 	&0.869 	&13.90		
    &\textbf{0.913} 	&0.775 	&\underline{10.50}
    &4.195 	&4.013 	&51.62	
    &-	&-	&-		
    &-	&-	&-		
    &\underline{0.921} 	&\textbf{0.663} 	&\textbf{10.14}		
    &1.266 	&\underline{0.685} 	&11.86\\

    & &20\% 
    &2.198 	&1.989 	&25.37		
    &1.720 	&1.042 	&15.40	
    &\textbf{1.116} 	&0.895 &\underline{11.90}	
    &3.517 	&3.302 	&41.95	
    &-	&-	&-		&-	&-	&-	
    &\underline{1.171} 	&\textbf{0.698} 	&\textbf{10.24}		
    &1.551 	&\underline{0.779} 	&12.16\\

    & &50\% 
    &1.695 	&1.457 	&18.46
    &1.997 	&1.191 	&16.40	
    &\textbf{1.167} 	&{0.898} 	&11.70
    &2.375 	&1.876 	&23.62	
    &-	&-	&-		&-	&-	&-	
    &\underline{1.404} 	&\textbf{0.758} 	&\textbf{10.49}		
    &1.594 	&\underline{0.778} 	&\underline{11.16}\\

    & &70\% 
    &1.527 	&1.279 	&16.20		
    &2.048 	&1.221 	&16.60	
    &\textbf{1.083}	&0.818 	&10.63	
    &2.009 	&1.370 	&17.25	
    &-	&-	&-		&-	&-	&-		
    &\underline{1.411} 	&\underline{0.771} 	&\underline{10.53}		&1.474 	&\textbf{0.709} 	&\textbf{10.02}\\

    & &80\% 
    &1.458 	&1.206 	&15.26	
    &2.063 	&1.230 	&16.70	
    &\textbf{1.038} 	&0.776 	&10.02	
    &1.881 	&1.230 	&15.46	
    &-	&-	&-		&-	&-	&-	
    &\underline{1.379} 	&\underline{0.764} 	&\underline{10.00}		
    &1.404 	&\textbf{0.668} 	&\textbf{9.40}\\

    & &100\% 
    &1.336 	&1.073 	&13.59	
    &2.085 	&1.243 	&16.80		
    &\textbf{0.973} 	&\underline{0.722} 	&\underline{9.300}	
    &1.693 	&1.059 	&13.28	
    &-	&-	&-		&-	&-	&-	
    &1.289 	&0.742 	&9.96		
    &\underline{1.274} 	&\textbf{0.592} 	&\textbf{8.27}\\
    \cmidrule{3-27}
    
    && Avg 
    &1.820 	&1.591 	&20.24 		
    &1.883 	&1.133 	&15.97 		
    &\textbf{1.048} 	&0.814 	&10.68	
    &2.612 	&2.142 	&27.20 										
    &-	&-	&-		&-	&-	&-	
    &\underline{1.263} 	&\underline{0.733} 	&\textbf{10.23} 		
    &1.427 	&\textbf{0.702} 	&\underline{10.48} \\

    \cmidrule{2-27}

    &\multirow{7}{*}{Heat}
    &10\% 
    &6.658 	&5.766 	&92.54		
    &0.162 	&0.120 	&2.310	
    &0.327 	&0.241 	&4.408		
    &5.880 	&4.877 	&74.50		
    &-	&-	&-		&-	&-	&-	
    &\underline{0.049} 	&\textbf{0.022} 	&\textbf{0.44}		
    &\textbf{0.038} 	&\underline{0.027} 	&\underline{0.51}\\
    
    & &20\% 
    &6.355 	&5.407 	&84.98		
    &0.203 	&0.154 	&3.000	
    &0.291 	&0.213 	&3.900	
    &5.725 	&4.700 	&71.50		
    &-	&-	&-		&-	&-	&-	
    &\underline{0.082} 	&\underline{0.048} 	&\underline{0.92}		
    &\textbf{0.039} 	&\textbf{0.028} 	&\textbf{0.53}\\
    
    & &50\%
    &5.462 	&4.360 	&69.19	
    &0.264 	&0.208 	&4.103	
    &0.227 	&0.162 	&3.018	
    &5.290 	&4.207 	&64.70		
    &-	&-	&-		&-	&-	&-	
    &\underline{0.095} 	&\underline{0.058} 	&\underline{1.13}		
    &\textbf{0.058} 	&\textbf{0.042} 	&\textbf{0.78}\\

    & &70\% 
    &5.007 	&3.881 	&66.50	
    &0.288 	&0.231 	&4.518	
    &0.204 	&0.147 	&2.709		
    &5.035 	&3.932 	&62.30	
    &-	&-	&-		&-	&-	&-	
    &\underline{0.086} 	&\underline{0.051} 	&\underline{0.99}		
    &\textbf{0.062} 	&\textbf{0.044} 	&\textbf{0.83}\\

    & &80\%
    &4.813 	&3.703 	&67.10	
    &0.298 	&0.240 	&4.715	
    &0.196 	&0.142 	&2.703	
    &4.907 	&3.800 	&61.50	
    &-	&-	&-		&-	&-	&-		
    &\underline{0.080} 	&\underline{0.046} 	&\underline{0.90}		
    &\textbf{0.061} 	&\textbf{0.044} 	&\textbf{0.83}\\

    & &100\% 
    &4.560 	&3.512 	&71.28		
    &0.313 	&0.254 	&4.904	
    &0.184 	&0.136 	&2.612		
    &4.687 	&3.583 	&61.10	
    &-	&-	&-		&-	&-	&-	
    &\underline{0.071} 	&\textbf{0.038} 	&\textbf{0.74}		
    &\textbf{0.059} 	&\underline{0.042} 	&\underline{0.80}\\
    \cmidrule{3-27}

    && Avg 
    &5.476 	&4.438 	&75.27 		
    &0.255 	&0.201 	&3.925 		
    &0.238 	&0.174 	&3.225 		
    &5.254 	&4.183 	&65.93 										
    &-	&-	&-		&-	&-	&-	
    &\underline{0.077} 	&\underline{0.044} 	&\underline{0.853} 		
    &\textbf{0.053} 	&\textbf{0.038} 	&\textbf{0.713} 	\\

    \cmidrule{2-27}
    
    &\multirow{6}{*}{2D-CFD }
    &10\% 	
    &0.347	&0.108	&30.70	
    &0.269 	&0.075 	&28.09	
    &\underline{0.060} 	&\textbf{0.017} 	&\underline{16.21}	
    &0.587	&0.218	&59.40	
    &-	&-	&-		&-	&-	&-		
    &\textbf{0.031} 	&\textbf{0.017} 	&\textbf{0.36}		
    &\textbf{0.031} 	&\underline{0.018} 	&\textbf{0.36}\\

    & &20\% 
    &0.342	&0.108	&29.60	
    &0.269 	&0.075 	&26.70	
    &\underline{0.060} 	&\underline{0.017} 	&\underline{15.60}	
    &0.574	&0.216	&58.70	
    &-	&-	&-		&-	&-	&-	
    &\textbf{0.030} 	&\textbf{0.016} 	&\textbf{0.34}		
    &\textbf{0.030} 	&\underline{0.017} 	&\textbf{0.34}\\
    
    & &50\% 
    &0.341	&0.107	&28.30	
    &0.268 	&0.074 	&25.30	
    &0.066 	&\underline{0.018} 	&\underline{14.10}		
    &0.573	&0.223	&56.30		
    &-	&-	&-		&-	&-	&-	
    &\underline{0.032} 	&\textbf{0.015} 	&\textbf{0.31}		
    &\textbf{0.031} 	&\textbf{0.015} 	&\textbf{0.31}\\
    
    & &70\% 
    &0.341	&0.108	&28.40	
    &0.267 	&0.074 	&24.80	
    &0.071 	&\underline{0.019} 	&13.30		
    &0.569	&0.236	&57.20	
    &-	&-	&-		&-	&-	&-	
    &\underline{0.036} 	&\textbf{0.015} 	&\underline{0.31}		
    &\textbf{0.034} 	&\textbf{0.015} 	&\textbf{0.30}\\

    & &80\% 	
    &0.343	&0.109	&26.30		
    &0.266 	&0.074 	&24.70	
    &0.075 	&\underline{0.020} 	&13.00	
    &0.572	&0.216	&56.10	
    &-	&-	&-		&-	&-	&-	
    &\underline{0.039} 	&\textbf{0.015} 	&\underline{0.31}		
    &\textbf{0.037} 	&\textbf{0.015} 	&\textbf{0.30}\\

    & &100\% 	
    &0.340	&0.110	&27.40	
    &0.265 	&0.073 	&24.40		
    &0.082 	&\underline{0.021} 	&12.50
    &0.570	&0.220	&56.90	
    &-	&-	&-		&-	&-	&-	
    &\underline{0.045} 	&\textbf{0.015} 	&\underline{0.32}		
    &\textbf{0.043} 	&\textbf{0.015} 	&\textbf{0.31}\\
    \cmidrule{3-27}

    && Avg 
    &0.342 	&0.108 	&28.45
    &0.267 	&0.074 	&25.67	
    &0.069 	&\underline{0.019} 	&14.12 		
    &0.574 	&0.222 	&57.43										
    &-	&-	&-		&-	&-	&-	
    &\underline{0.036} 	&\textbf{0.016} 	&\underline{0.325} 		
    &\textbf{0.034} 	&\textbf{0.016} 	&\textbf{0.320} 	\\
    \cmidrule{2-27}
        
    &\multirow{6}{*}{Gene}
    &10\% 
    &0.722 	&0.531 	&55.50	
    &\underline{0.502}	&\textbf{0.409}	&\underline{42.60}	
    &0.563 	&0.415 	&45.30		
    &1.301	&1.192	&75.68	
    &-	&-	&-		&-	&-	&-	
    &0.701 	&0.578 	&\textbf{40.19}		
    &\textbf{0.480} 	&\underline{0.410} 	&47.32\\
    
    & &20\% 
    &0.667 	&0.492 	&48.40	
    &\underline{0.551}	&\underline{0.452}	&47.90		
    &0.593 	&0.454 	&47.80	
    &1.233	&1.135	&67.63	
    &-	&-	&-		&-	&-	&-	
    &0.704 	&0.585 	&\textbf{37.86}		
    &\textbf{0.479} 	&\textbf{0.401} 	&\underline{40.50}\\

    & &50\% 	
    &\underline{0.568} 	&\underline{0.406} 	&\underline{37.90}		
    &0.687	&0.565	&53.90		
    &0.703 	&0.564 	&53.20	
    &1.068	&0.988	&71.66	
    &-	&-	&-		&-	&-	&-	
    &0.740 	&0.621 	&38.10		
    &\textbf{0.515} 	&\textbf{0.397} 	&\textbf{33.15}\\

    & &70\%
    &\underline{0.564} 	&\textbf{0.388} 	&\underline{36.00}	
    &0.744	&0.611	&48.30
    &0.779 	&0.628 	&54.10		
    &0.981	&0.903	&68.95	
    &-	&-	&-		&-	&-	&-	
    &0.780 	&0.652 	&39.49		
    &\textbf{0.558} 	&\underline{0.413} 	&\textbf{31.61}\\

    & &80\% 
    &\textbf{0.576} 	&\textbf{0.389} 	&\underline{35.60}		
    &0.765	&0.629	&52.30	
    &0.816 	&0.657 	&54.50	
    &0.944	&0.864	&66.31	
    &-	&-	&-		&-	&-	&-	
    &0.801 	&0.668 	&40.02		
    &\underline{0.580} 	&\underline{0.423} 	&\textbf{31.25}\\

    & &100\% 
    &\textbf{0.619} 	&\textbf{0.401} 	&\underline{35.30}	
    &0.799	&0.658	&51.70	
    &0.884 	&0.710 	&54.90	
    &0.880	&0.790	&59.94	
    &-	&-	&-		&-	&-	&-
    &0.841 	&0.696 	&40.53		
    &\underline{0.623} 	&\underline{0.448} 	&\textbf{30.58}\\
    \cmidrule{3-27}

    && Avg
    &\underline{0.619} 	&\underline{0.435} 	&41.45		
    &0.675 	&0.554 	&49.45		
    &0.723 	&0.571 	&51.63 		
    &1.068 	&0.979 	&68.36										
    &-	&-	&-		&-	&-	&-
    &0.761 	&0.633 	&\underline{39.37} 		
    &\textbf{0.539} 	&\textbf{0.415} 	&\textbf{35.74} 	\\

    \midrule

    \multirow{8}{*}{\rotatebox{90}{Real-world Systems}}
    &T-Drive	&12	
    &282.1 	&149.5 	&77.91	
    &102.7 	&52.36 	&110.1
    &132.1 	&69.28 	&89.40		
    &144.0 	&78.88 	&87.36	
    &\textbf{51.57} 	&25.18 	&41.10	
    &269.8 	&143.8 	&87.60	
    &\underline{56.38} 	&\textbf{22.93} 	&\underline{31.27}		
    &60.33 	&\underline{24.38} 	&\textbf{30.61}\\

    &NYCTaxi	&12	
    &197.2 	&122.1  &87.00		
    &53.22 	&28.85 	&62.60		
    &61.74 	&34.44 	&79.80		
    &130.0  &69.27 	&88.46	
    &38.38 	&21.08 	&48.00	
    &102.5  &45.65 	&48.10	
    &\underline{31.19}  &\underline{16.73}  &\underline{33.52}
    &\textbf{27.67}  &\textbf{15.31}  &\textbf{31.26}\\

    &CHIBike	&12	
    &\textbf{4.365}  &\textbf{0.759} 	&\textbf{14.10}
    &8.111 	&4.275 	&91.90	
    &10.04 	&5.167 	&89.40	
    &13.48 	&6.396 	&\underline{55.10}
    &\underline{5.040}  &\underline{2.768} 	&61.40		
    &11.86 	&6.013 	&87.40	
    &5.806 	&3.127 	&70.22		
    &5.108 	&2.874 	&67.03\\
    												
    &PEMS03	&12
    &77.70 	&45.07  &41.89	
    &52.26 	&33.37  &41.30	
    &55.26 	&35.63  &44.20	
    &218.9  &167.7  &94.00	
    &31.36 	&18.04  &19.90	
    &\underline{23.83} 	&\textbf{13.29}  &\textbf{14.46}
    &\textbf{23.40}  &\underline{14.60}  &\underline{14.58}		
    &24.59  &15.55  &16.83\\

    &PEMS04	&12	
    &54.63  &34.02 	&22.00	
    &53.71 	&36.40 	&30.90		
    &57.96 	&39.66 	&31.00		
    &254.0  &200.3 	&86.60	
    &31.33 	&19.98 	&15.02
    &125.9  &70.18 	&35.40		
    &\textbf{28.22}  &\textbf{17.61} 	&\textbf{12.87}		
    &\underline{30.59} 	&\underline{19.32} 	&\underline{14.78}\\
    	
    &PEMS07	&12
    &76.37 	&54.25 	&32.40		
    &81.49 	&58.61	&36.30		
    &83.74 	&57.27 	&30.60	
    &323.1  &266.8  &88.50	
    &39.59 	&25.78 	&11.90	
    &\underline{35.46} 	&23.13 	&\textbf{9.700}
    &\textbf{31.67}  &\textbf{19.92}  &\underline{10.99}		
    &36.29  &\underline{23.01}  &12.73\\
    										
    &PEMS08	&12
    &47.13 	&29.73 	&16.90	
    &43.84 	&29.39 	&24.30	
    &45.88 	&30.56 	&24.50	
    &258.3  &213.6  &93.60	
    &26.98 	&17.25 	&11.50
    &129.5  &70.22 	&32.60		
    &\textbf{21.23}  &\textbf{13.16} &\textbf{9.05}		
    &\underline{23.16}  &\underline{14.38} &\underline{11.12}\\

    &NOAA	&12	
    &10.78 	&8.482  &61.30	
    &12.99 	&10.05 	&70.50		
    &10.87 	&8.222  &49.00	
    &30.36 	&25.71 	&84.90	
    &8.160 	&5.994  &59.70	
    &35.00 	&28.39 	&83.80	
    &\textbf{6.646} 	&\textbf{4.736} 	&\textbf{21.50}	
    &\underline{6.965} 	&\underline{4.977} 	&\underline{22.71}\\

    \bottomrule
    \end{tabular}}
    \label{T_fore_full}
% \end{sidewaystable*}
\end{table*}

\textit{Fine-tuning Settings}: For efficiency, we randomly select $M_s=\{1500, 1500, 200, 50\}$ samples for Mutualistic, Heat, 2D-CFD and Gene as fine-tuning dataset. For synthetic systems, each observation sequence is split into two sub-sequences as training and test sample. The training sequence length is set as 180 for Mutualistic and 120 for others. To form short/long-term forecasting, during testing, we first forecast the whole test sequence and then truncate the test sequence by ratios of $\{10\%, 20\%, 50\%, 70\%, 80\%, 100\%\}$ to form 6 sub-sequences and calculate metrics for each of them. For real-world systems, we adopt the same train/test sets splitting ratios and forecasting settings following \cite{STGODE2021}: We split the overlong observation sequence by look-back window of length 24 and stride 1. The samples are split into train/valid/test sets by ratios of 60\%, 20\% and 20\%. The history and forecasting lengths are both set as 12, \ie using the past 12 time points to forecast the future 12 points. 
The learning rates are tuned over $\ 1e-4\sim1e-6$. The patch length and stride are tuned over $ \{12,16,20,24,30,36\}$ and $ \{2,4,6,8,10\}$, respectively. The first and last patch of each training sequence are adopted as the initial values for training and testing, respectively.

\begin{figure}[!t]
    \centering
    \includegraphics[width=0.9\linewidth]{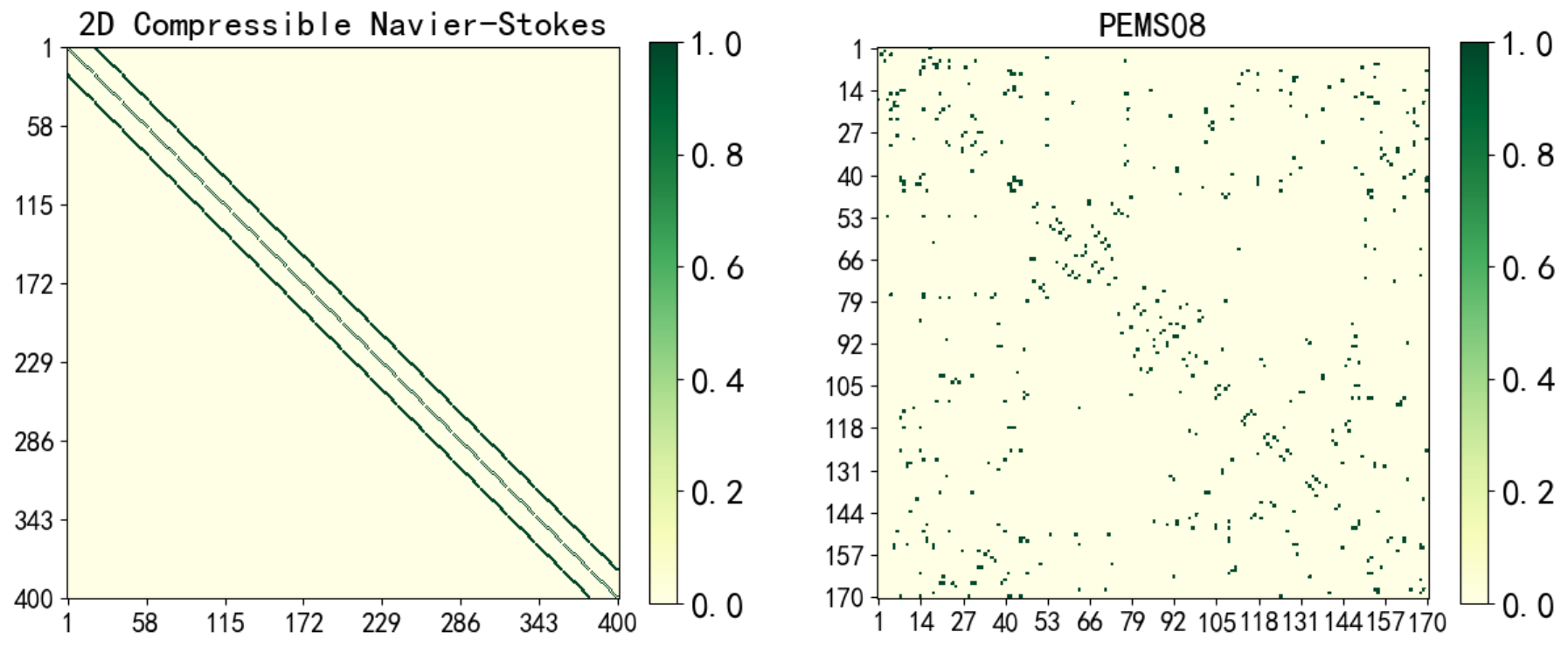} 
    \caption{The interacting graphs among objects of 2D Compressible Navier-Stokes (2D-CFD) and PEMS08.}
    \label{fg_corr}
\end{figure}

\begin{table*}[!t]
    \centering
    \setlength{\tabcolsep}{0.3pt}
    % \scriptsize
    \tiny	
    \caption{Full results of cross-domain settings and pre-training variants on \baby-small. Each score is the averaged results of multiple sample with different initial values. The best scores are in \textbf{boldface} and the second best scores are in \underline{underline}. ``Forecast Length'' denotes the forecasting length. For synthetic systems, we truncate the test sequence by ratios of $\{10\%,20\%,50\%,70\%,80\%,100\%\}$ to form short/long-term forecasting. ``Avg'' denotes the averaged results of all forecasting lengths. MAPE is in \%. ``\baby-para'' denotes the variant fine-tuning under the cross-domain setting of pre-training learning one parameter out; ``\baby-sys'' denotes the variant fine-tuning under the cross-domain setting of pre-training leaving one system out; ``\baby w/o MLE'' denotes the ablative version pre-training without maximal Lyapunov Exponent objective;  ``\baby w/o pre'' denotes the variant which fine-tunes without pre-training; ``\baby frz'' denotes the variant which fine-tunes by freezing our pre-trained PLM module. ``-'' denotes there are no available system-specific parameters.}
    \renewcommand\arraystretch{1.0}
    \resizebox{\linewidth}{!}{
    % \begin{tabular}{p{40pt}<{\centering}| p{25pt}<{\centering} |
    % p{20pt}<{\centering} p{20pt}<{\centering} p{20pt}<{\centering} |
    % p{20pt}<{\centering} p{20pt}<{\centering} p{20pt}<{\centering} |
    % p{20pt}<{\centering} p{20pt}<{\centering} p{20pt}<{\centering} |
    % p{20pt}<{\centering} p{20pt}<{\centering} p{20pt}<{\centering} |
    % p{20pt}<{\centering} p{20pt}<{\centering} p{20pt}<{\centering} |
    % p{20pt}<{\centering} p{20pt}<{\centering} p{20pt}<{\centering} }

    \begin{tabular}{p{17pt}<{\centering} p{40pt}<{\centering}|  p{22pt}<{\centering} |
    % p{18pt}<{\centering}  p{18pt}<{\centering}  p{18pt}<{\centering} | 
    p{18pt}<{\centering}  p{18pt}<{\centering}  p{18pt}<{\centering} | 
    p{18pt}<{\centering}  p{18pt}<{\centering}  p{18pt}<{\centering} | 
    p{18pt}<{\centering}  p{18pt}<{\centering}  p{18pt}<{\centering} | 
    p{18pt}<{\centering}  p{18pt}<{\centering}  p{18pt}<{\centering} | 
    p{18pt}<{\centering}  p{18pt}<{\centering}  p{18pt}<{\centering} | 
    p{18pt}<{\centering}  p{18pt}<{\centering}  p{18pt}<{\centering} 
    }
    \toprule
    
    \multicolumn{2}{c|}{\multirow{2}{*}{System}}   
    & \multirow{2}{*}{\makecell{Forecast\\Length}} 
    &\multicolumn{3}{c|}{\baby-small}       
    &\multicolumn{3}{c|}{\baby-para}     
    &\multicolumn{3}{c|}{\baby-sys}
    &\multicolumn{3}{c|}{\baby w/o MLE}       
    &\multicolumn{3}{c|}{\baby w/o pre}
    &\multicolumn{3}{c}{\baby frz}\\
    & & &RMSE &MAE &MAPE
    &RMSE &MAE &MAPE
    &RMSE &MAE &MAPE
    &RMSE &MAE &MAPE
    &RMSE &MAE &MAPE
    &RMSE &MAE &MAPE\\
    
    \midrule																
    \multirow{29}{*}{\rotatebox{90}{Synthetic Systems}}
    &\multirow{7}{*}{Mutualistic}									
	&10\%	&\textbf{1.266}	&\textbf{0.685}	&\textbf{11.86}		&1.461	&0.734	&13.92		&2.573	&2.293	&30.19		&1.443	&1.067	&15.14		&1.892	&1.377	&20.39		&4.438	&4.339	&57.54		\\
	&&20\%	&\textbf{1.551}	&\textbf{0.779}	&\textbf{12.16}		&1.682	&0.792	&14.85		&2.709	&2.417	&31.21		&1.694	&1.228	&16.74		&1.974	&1.358	&19.27		&4.595	&4.505	&58.36		\\
	&&50\%	&\textbf{1.594}	&\textbf{0.778}	&\textbf{11.16}		&1.634	&0.801	&14.46		&2.688	&2.367	&30.03		&1.830	&1.313	&17.29		&1.807	&1.180	&16.08		&4.648	&4.563	&57.9		\\
	&&70\%	&\textbf{1.474}	&\textbf{0.709}	&\textbf{10.02}		&1.602	&0.797	&14.07		&2.530	&2.177	&27.58		&1.788	&1.311	&17.12		&1.622	&1.036	&13.99		&4.525	&4.431	&56.04		\\
	&&80\%	&\textbf{1.404}	&\textbf{0.668}	&\textbf{9.40}		&1.593	&0.763	&13.46		&2.421	&2.051	&25.99		&1.756	&1.315	&17.09		&1.536	&0.970	&13.06		&4.401	&4.282	&54.15		\\
	&&100\%	&\textbf{1.274}	&\textbf{0.592}	&\textbf{8.27}		&1.483	&0.77	&13.17		&2.237	&1.871	&23.66		&1.705	&1.333	&17.17		&1.392	&0.858	&11.49		&4.055	&3.789	&47.99		\\
    \cmidrule{3-21}		
	&&Avg	&\textbf{1.427}	&\textbf{0.702}	&\textbf{10.48}		&1.576	&0.776	&13.99		&2.526	&2.196	&28.11		&1.702	&1.261	&16.76		&1.704	&1.130	&15.71		&4.444	&4.318	&55.33		\\

    \cmidrule{2-21}																
    																
    & \multirow{7}{*}{Heat}														
	&10\%	&\textbf{0.038} 	&\textbf{0.027} 	&\textbf{0.51}		&0.041	&0.032	&0.57		&0.061	&0.044	&0.82		&{0.059} 	&{0.051} 	&{0.98}		&0.062	&0.048	&0.91		&0.063	&0.045	&0.87		\\
	&&20\%	&\textbf{0.039} 	&\textbf{0.028} 	&\textbf{0.53}		&0.068	&0.049	&0.89		&0.074	&0.053	&0.99		&0.056 	&0.045 	&0.87		&0.074	&0.055	&1.02		&0.07	&0.051	&0.97		\\
	&&50\%	&\textbf{0.058} 	&\textbf{0.042} 	&\textbf{0.78}		&0.097	&0.072	&1.25		&0.100	&0.074	&1.38		&0.067 	&0.050 	&0.95		&0.099	&0.072	&1.34		&{0.078}	&{0.058}	&{1.09}		\\
	&&70\%	&\textbf{0.062} 	&\textbf{0.044} 	&\textbf{0.83}		&0.099	&0.071	&1.27		&0.098	&0.073	&1.37		&0.066 	&0.050 	&0.94		&0.101	&0.075	&1.39		&{0.078}	&{0.058}	&{1.09}		\\
	&&80\%	&\textbf{0.061} 	&\textbf{0.044} 	&\textbf{0.83}		&0.091	&0.071	&1.30		&0.095	&0.071	&1.33		&0.065 	&0.049 	&0.92		&0.099	&0.073	&1.37		&{0.077}	&{0.058}	&{1.08}		\\
	&&100\%	&\textbf{0.059} 	&\textbf{0.042} 	&\textbf{0.80}		&0.090	&0.064	&1.19		&0.088	&0.065	&1.22		&0.062 	&0.046 	&0.86		&0.094	&0.068	&1.28		&{0.075}	&{0.055}	&{1.04}		\\
    \cmidrule{3-21}		
	&&Avg	&\textbf{0.053} 	&\textbf{0.038} 	&\textbf{0.71}		&0.089	&0.060	&1.08		&0.086	&0.063	&1.18		&0.063 	&0.049 	&0.92		&\textbf{0.088}	&\textbf{0.065}	&1.22		&{0.073}	&{0.054}	&1.02		\\
    \cmidrule{2-21}																
    																
    &\multirow{7}{*}{2D-CFD}							
	&10\%	&0.031	&0.018	&0.36		&0.034	&\textbf{0.017}	&\textbf{0.35}		&0.03	&0.018	&0.37		&\textbf{0.028}	&\textbf{0.017}	&0.36		&0.030	&\textbf{0.017}	&0.36		&0.052	&0.024	&0.52		\\
	&&20\%	&0.03	&0.017	&\textbf{0.34}		&0.032	&0.018	&0.37		&0.03	&0.017	&0.36		&\textbf{0.027}	&0.017	&\textbf{0.34}		&\textbf{0.030}	&\textbf{0.016}	&\textbf{0.34}		&0.055	&0.025	&0.53		\\
	&&50\%	&0.031	&\textbf{0.015}	&\textbf{0.31}		&0.034	&0.018	&0.35		&0.034	&0.017	&0.34		&\textbf{0.03}	&0.016	&0.32		&0.034	&0.016	&0.33		&0.063	&0.026	&0.55		\\
	&&70\%	&\textbf{0.034}	&\textbf{0.015}	&\textbf{0.30}		&0.039	&0.016	&0.33		&0.04	&0.017	&0.34		&0.036	&0.016	&0.32		&0.041	&0.016	&0.34		&0.07	&0.027	&0.57		\\
	&&80\%	&\textbf{0.037}	&\textbf{0.015}	&\textbf{0.30}		&0.041	&0.017	&0.33		&0.044	&0.017	&0.35		&0.039	&0.016	&0.32		&0.045	&0.016	&0.34		&0.074	&0.027	&0.58		\\
	&&100\%	&0.043	&\textbf{0.015}	&\textbf{0.31}		&\textbf{0.037}	&0.016	&0.35		&0.052	&0.017	&0.36		&0.047	&0.016	&0.33		&0.053	&0.017	&0.36		&0.083	&0.028	&0.61		\\
    \cmidrule{3-21}		
	&&Avg	&\textbf{0.034}	&\textbf{0.016}	&\textbf{0.32}		&0.036	&0.17	&0.35		&0.038	&0.017	&0.35		&0.035	&\textbf{0.016}	&0.33		&0.039	&\textbf{0.016}	&0.35		&0.066	&0.026	&0.56		\\
    \cmidrule{2-21}

    &\multirow{7}{*}{Gene}							
	&10\%	&0.48	&0.41	&47.32		&0.725	&0.603	&49.32		&0.756	&0.655	&51.58		&\textbf{0.415}	&\textbf{0.32}	&\textbf{33.84}		&1.117	&0.901	&100.72		&0.866	&0.782	&63.96		\\
	&&20\%	&0.479	&0.401	&40.50		&0.734	&0.597	&47.96		&0.743	&0.649	&48.38		&\textbf{0.441}	&\textbf{0.325}	&\textbf{32.72}		&1.006	&0.817	&89.26		&0.911	&0.826	&67.64		\\
	&&50\%	&\textbf{0.515}	&\textbf{0.397}	&33.15		&0.782	&0.583	&38.75		&0.693	&0.604	&43.64		&0.56	&0.413	&\textbf{32.39}		&0.865	&0.704	&71.98		&1.045	&0.955	&74.82		\\
	&&70\%	&\textbf{0.558}	&\textbf{0.413}	&\textbf{31.61}		&\textbf{0.803}	&\textbf{0.549}	&34.92		&0.672	&0.576	&40.87		&0.646	&0.488	&32.71		&0.856	&0.692	&64.72		&1.13	&1.033	&76.35		\\
	&&80\%	&\textbf{0.58}	&\textbf{0.423}	&\textbf{31.25}		&0.794	&0.557	&35.37		&0.668	&0.565	&39.67		&0.686	&0.523	&32.86		&0.862	&0.694	&61.96		&1.17	&1.069	&76.6		\\
	&&100\%	&\textbf{0.623}	&\textbf{0.448}	&\textbf{30.58}		&0.801	&0.541	&35.83		&0.667	&0.549	&37.48		&0.756	&0.586	&33.11		&0.882	&0.704	&57.2		&1.244	&1.136	&76.26		\\
    \cmidrule{3-21}		
	&&Avg	&\textbf{0.539}	&\textbf{0.415}	&35.73		&0.773	&0.572	&40.36		&0.700	&0.600	&43.6		&0.584	&0.443	&\textbf{32.94}		&0.931	&0.752	&74.31		&1.061	&0.967	&72.61		\\

    \midrule																
    
    \multirow{8}{*}{\rotatebox{90}{Real-world Systems}}
    &T-Drive	&12	&60.33	&24.384	&\textbf{30.61}		&-	&-	&-		&\textbf{56.49}	&\textbf{24.28}	&30.61		&72.07	&32.64	&38.32		&67.86	&30.9	&37.06		&304.9	&169.2	&90.57		\\
    
    &CHIBike	&12	&5.108	&2.874	&67.03		&-	&-	&-		&4.878	&\textbf{2.791}	&66.25		&4.976	&2.835	&\textbf{66.5}		&\textbf{4.838}	&2.799	&68.15		&9.788	&4.408	&70.81		\\
    
    &NYCTaxi	&12	&\textbf{0.004}	&\textbf{0.002}	&\textbf{32.50}		&-	&-	&-		&\textbf{0.004}	&\textbf{0.002}	&36.49		&\textbf{0.004}	&\textbf{0.002}	&35.81		&0.005	&0.003	&49.82		&0.009	&0.004	&58.5		\\
    
    &PEMS03	&12	&\textbf{24.59}	&\textbf{15.55}	&\textbf{16.83}		&-	&-	&-		&27.42	&17.20	&17.9		&27.66	&17.50	&18.51		&29.26	&18.61	&21.40		&53.52	&33.75	&81.42		\\
    
    &PEMS04	&12	&\textbf{30.60}	&\textbf{19.32}	&14.78		&-	&-	&-		&31.23	&19.90	&\textbf{14.72}		&31.71	&20.69	&17.06		&34.75	&22.32	&18.56		&68.60	&48.14	&80.63		\\
    
    &PEMS07	&12	&\textbf{36.29}	&\textbf{23.01}	&\textbf{12.73}		&-	&-	&-		&162.0	&130.8	&193.2		&40.56	&26.53	&13.48		&42.12	&27.67	&15.35		&96.89	&69.34	&118.4		\\
    
    &PEMS08	&12	&23.16	&14.38	&\textbf{11.12}		&-	&-	&-		&\textbf{22.96}	&\textbf{14.16}	&11.71		&23.25	&14.58	&11.81		&24.16	&14.99	&12.75		&43.56	&28.99	&47.73		\\
    
    &NOAA	&12	&6.965	&4.977	&22.71		&-	&-	&-		&\textbf{6.376}	&\textbf{4.623}	&\textbf{20.41}		&7.195	&5.17	&23.73		&7.375	&5.334	&24.68		&9.274	&7.086	&35.61		\\

    \bottomrule
    % \Xhline{1.5pt}
    \end{tabular}}
    \label{T_abla_full}
\end{table*}

\paragraph{Calculation of Interacting Graphs} 
For Mutualistic, Heat, 2D-CFD, Gene, we adopt a grid Network as the interacting graph following \cite{NDCN2020}, where each object interacts with its 8 neighbors.
% For NYCTaxi ,T-Drive, CHIBike and NOAA, the interaction graphs are distances between each station (\ie object) calculated by the geometry coordinates or grid network.
% For PEMS03, PEMS04, PEMS07 and PEMS08, we calculate the interacting graph following \cite{STGODE2021} as below:
For real-world systems, we calculate the interacting graph following \cite{STGODE2021} as below:
\begin{equation}
    \mathbf{A}_{ij} = \left\{ 
                            \begin{aligned}
                                &\exp\left( \frac{{\rm Dst}(i,j)^2}{\delta^2} \right),\ \  {\rm if} \;\exp\left( \frac{{\rm Dst}(i,j)^2}{\delta^2} \right) \geq \epsilon \\
                              &0,  \, \ \  \quad \quad\quad \quad\quad \quad\quad  {\rm otherwise} 
                            \end{aligned}
                        \right.,\nonumber
\end{equation}
where ${\rm Dis}(i,j)$ denotes the distance between object $i$ an $j$; $\delta^2$ and $\epsilon$ are the sparsity controlling thresholds. For NYCTaxi, T-Drive, CHIBike and NOAA, the distances are calculated by the geometry coordinates or grid network. For PEMS03, PEMS04, PEMS07 and PEMS08, the distances are built-in from the original datasets. We present the interacting graphs of 2D-CFD and PEMS08 in Fig. \ref{fg_corr}.

\subsection{Dynamics Modeling Results}

% \textit{Evaluation Metrics}: 
We adopt RMSE, MAE and MAPE (in \%) for evaluating the short/long-term forecasting performance. Specially, each result is the averaged scores of multiple samples with \textit{different initial values} for identical system. 
% and report full and averaged results in Table \ref{T_fore_full}. 

% \subsubsection{Interpolation Task}

% \subsubsection{Extrapolation Task}

\paragraph{In-domain Forecasting} 
% one dynamics, different hyper-parameters for pre-training and fine-tuning
% one dynamics, first half sequence for pre-training, and the rest half for fine-tuning

We first examine \baby by short/long-term forecasting on in-domain settings against 6 baseline methods. We pre-train \baby on all 152 sets of observations and fine-tune on 1 set of observations for each system (consisting of multiple samples generated by an identical system, each sample corresponds with different initial values).
The full forecasting results are presented in Table \ref{T_fore_full}. Overall speaking, our \baby outperforms baseline methods in most settings and gain significant improvements over baselines. For example, the performance gaps on RMSE and MAE are about 3.98 and 4.09 for PEMS08, 0.185 and 0.136 for Heat Diffusion. More importantly, the significant improvements on MAPE directly indicating that our \baby can perform more efficient and well-fitted forecasting sequence. For example, the performance gaps are about 14\% and 17\% on MAPE for 2D-CFD and NYCTaxi. 
Besides, We observe that \baby-small outperforms \baby on synthetic systems, especially for the average results, whereas the opposite holds on real-world systems. This phenomenon suggests that a smaller backbone may be more efficient in handling cleaner datasets, yielding more stable and robust performance, whereas a larger backbone is better suited for capturing complex temporal patterns in non-stationary and noisy real-world observations. In a nutshell, our \baby can efficiently approximate hidden dynamics and perform well in both short/long-term forecasting.  

\paragraph{Cross-domain Forecasting} 
% Table \ref{T_crossdomain}
% Leave-one-out : 1. systems 2. args
% short/long term
% The two variants are denoted as ``\baby-sys'' and ``\baby-para''. 
We examine the generalizability of our \baby on cross-domain settings with backbone of {\ttfamily{T5-small}}. We set two Leave-One-Out (LOO) cross-domain settings, leaving one system out and leaving one set of parameters out. (1) \textbf{\baby-para}: For LOO on hyper-parameters, we pre-train \baby on observations excluding observations of a specific parameter set $\mathcal{G}_m$ and fine-tune on $\mathcal{G}_m$ for examination. (2) \textbf{\baby-sys}: For LOO on system $s$, we pre-train \baby on observations excluding all sets of observations of system $s$ and fine-tune on $\mathcal{G}_m$ for examination. We conduct same forecasting examination as in-domain setting and present results in Table \ref{T_abla_full}. 
We can find that, the performance of in-domain setting outperforms the cross-domain settings in most cases. Meanwhile, the performance of excluding one system out also beats the in-domain setting in some cases and the overall performance gaps are not too large. 
% These phenomena indicate the strong generalizability of our \baby. 
Even pre-training under cross-domain settings, the variant of \baby is also on a par with in-domain settings. These phenomena indicate that our \baby exhibits strong generalization capacity and can serve as an effective dynamics embedder, thereby facilitating dynamics modeling for previously unseen systems.

\paragraph{Pre-training Variants and Ablative Study}
We examine the impact of pre-training on downstream dynamics modeling by modifying the initializations of the encoder and decoder when fine-tuning and setting an ablative objective version which pre-trains \baby without the M-Lyapunov Exponent objective.
The ablative version and variants on \baby-small include: (1) \textbf{\baby w/o MLE}: pre-train \baby without the Lyapunov objective, only with reconstruction and forecasting objectives; (2) \textbf{\baby w/o pre}: initialize by PLM, \ie fine-tune \baby without pre-training; (3) \textbf{\baby frz}: initialize by pre-trained \baby and fine-tuning \baby by freezing the pre-trained encoder and decoder.
The full results are presented in Table \ref{T_abla_full}. We can find that the full version with pre-training \baby consistently outperforms the variants, indicating the benefit of our pre-training on massive dynamics observations to learn hidden dynamics in latent space. Besides, we surprisingly find that the variant freezing the pre-trained encoder and decoder outperforms the variant without pre-training in some settings. These phenomena indicate that the pre-training processes can effectively capture the dynamics properties, leading to less efforts on fine-tuning processes when learning specific dynamics. We also find that the ablative version pre-training without Lyapunov objective outperforms other settings in some of the short-term forecasting settings on 2D-CFD and Gene. These phenomena may indicate that reducing the chaos from embedded observations when pre-training are more likely to benefit long-term forecasting to some extent. According to these results, our \baby can be directly adopted as an effective embedder on learning specific dynamics in real-world applications.
% when fine-tuning are unavailable.

% to examine whether \baby works when learning specific dynamics.
% We set two ablative versions, (1) Fine-tune \baby without pre-training, denoted as ``\baby w/o pre''; (2) Fine-tuning \baby with freezing the PLM encoder/decoders, denoted as ``\baby freeze $\bm\Theta$''. The average results are presented in Table \ref{T_ablative} and full results are presented in Appendix \ref{App_full_indomain}.

% \paragraph{Learn dynamics with white-box SINDy}
% a white-box dynamics leaner SINDy \cite{VAESINDY2019}
% To explicitly extract the analytical solution of the hidden dynamics, we adopted SINDy Auto-Encoders \cite{VAESINDY2019} as the dynamics learner, which imposes the white-box sparsity regression dynamics learner \cite{SINDY2016} on the original observations and latent representations simultaneously.

\begin{figure*}[!t]
    \centering
    \includegraphics[width=0.9\linewidth]{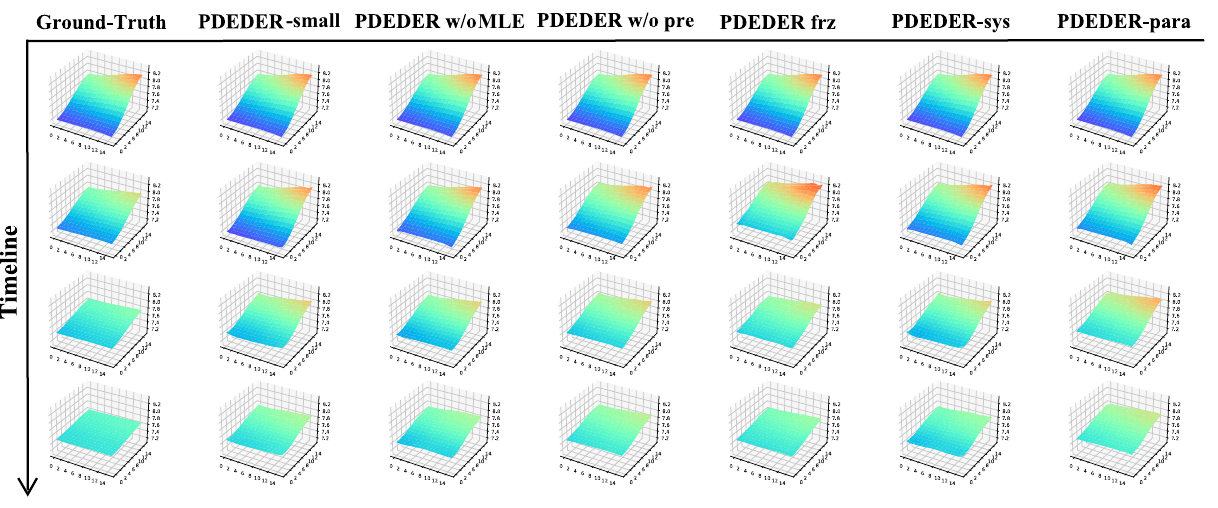} 
    \caption{Forecasting visualizations on Heat dynamics over the variants of \baby.  ``\baby w/o MLE'' denotes the ablative version pre-training without M-Lyapunov exponent; ``\baby w/o pre'' denotes fine-tuning \baby without pre-training; ``\baby frz'' denotes fine-tuning by freezing the PLM module; ``\baby-sys'' denotes the variant fine-tuning under the cross-domain setting of pre-training leaving one system out; ``\baby-para'' denotes the variant fine-tuning under the cross-domain setting of pre-training learning one parameter out. The horizontal axes denote object indices, while the vertical axis represents object state values.}
    \label{fg_vis_abla}
\end{figure*}

\begin{figure*}[ht]
% \begin{figure}[!t]
\centering
	% \subfloat[CHIBike]{\includegraphics[width = 0.45\textwidth]{plot/vis_CHIBike.pdf}} \hspace{5mm}
	% \hfill
    % \hspace{10mm} 
	\subfloat[T-Drive]{\includegraphics[width = 0.45\textwidth]{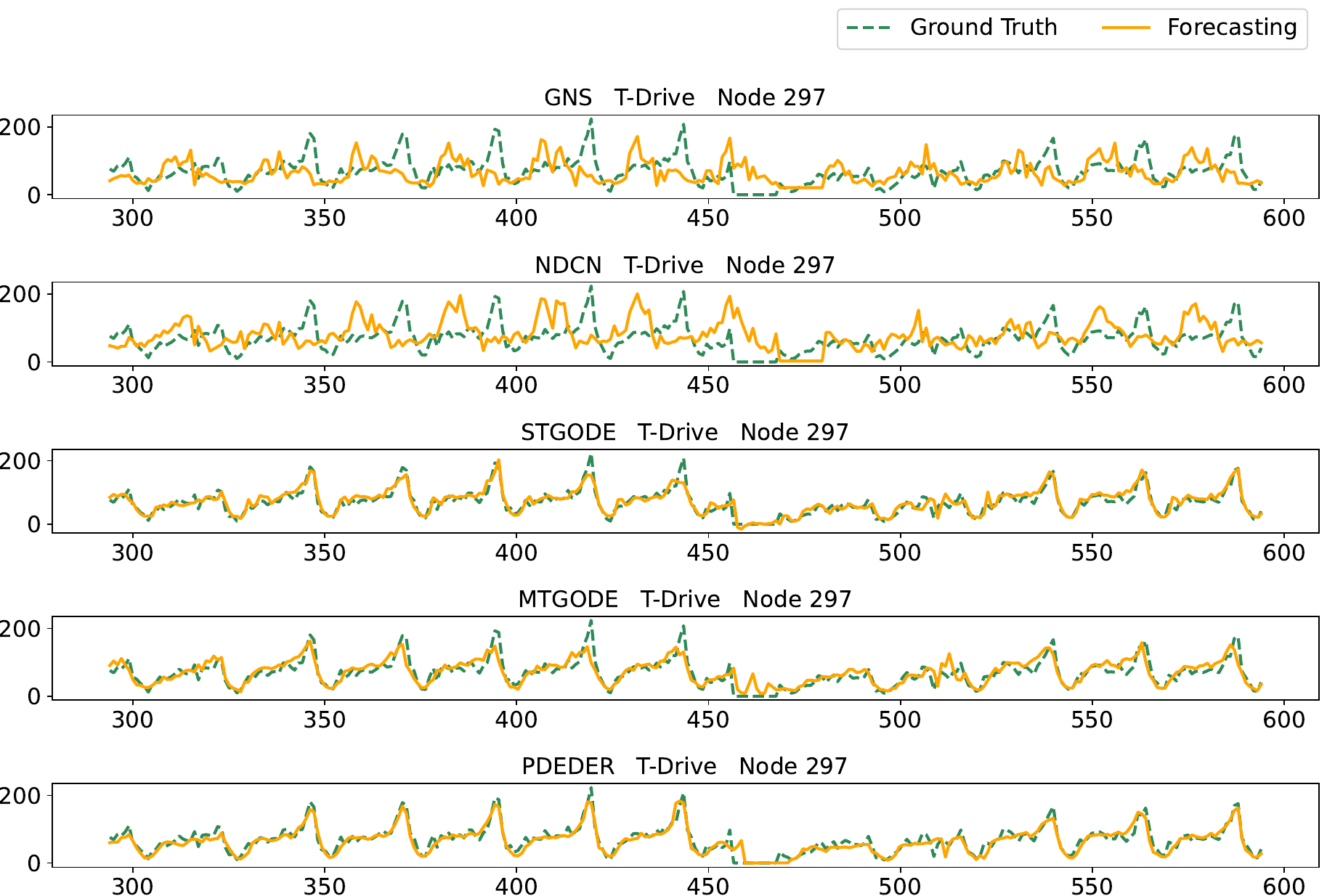}}
    \hspace{0.75cm}
    % \hspace{100mm} 
	% \hfill
	% \subfloat[PEMS03]{\includegraphics[width = 0.45\textwidth]{plot/vis_PEMS03.pdf}} \hspace{5mm}
    % \hfill
    % \hspace{10mm} 
	\subfloat[PEMS08]{\includegraphics[width = 0.45\textwidth]{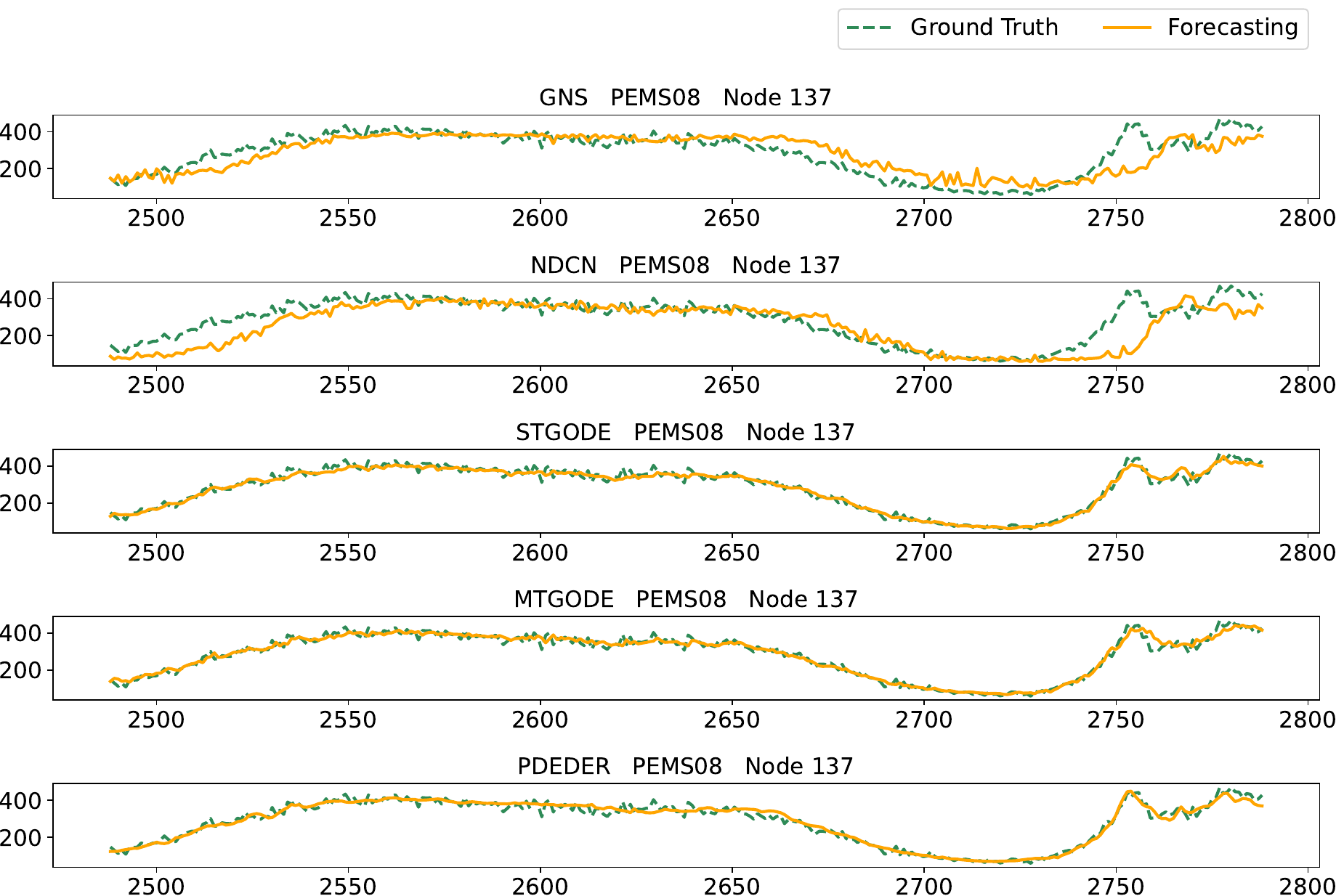}} 
% \caption{Traffic flow forecasting visualizations between \baby and baselines on real-world systems (a) CHIBike, (b) T-Drive, (c) PEMS03, (d) PEMS08. The vertical coordinate denotes the traffic flow values.}
\caption{Traffic flow forecasting visualizations on \baby and baselines of real-world systems (a) T-Drive and (b) PEMS08. The vertical coordinate denotes the traffic flow values.}
\label{fg_vis_real}
\end{figure*}

\paragraph{Forecasting Visualization} 
We present two forecasting visualizations tasks on synthetic and real-world systems. For synthetic systems, we present 4 forecasting visualization shots for \baby-small along with its all variants and ablative version mentioned above in Fig. \ref{fg_vis_abla}. For real-world system, we present forecasting comparisons against baseline methods on T-Drive and PEMS08 in Fig. \ref{fg_vis_real}. We randomly select 2 object nodes and a random time series of length 300. We can observe that \baby performs comparable dynamics against the ground-truth values and smoother evolving behaviors for both synthetic and real-world systems. Specially, our \baby can generate sequences with less delay on real-world systems comparing with baselines and fit the ground-truth sequences better.

% \end{figure}

% % \begin{figure*}[ht]
% \begin{figure*}[!t]
% \centering
% 	\subfloat[CHIBike]{\includegraphics[width = 0.45\textwidth]{plot/vis_CHIBike.pdf}} \hspace{5mm}
% 	% \hfill
%     % \hspace{10mm} 
% 	\subfloat[T-Drive]{\includegraphics[width = 0.45\textwidth]{plot/vis_TDrive.pdf}}
%     % \hspace{100mm} 
% 	\hfill
% 	\subfloat[PEMS03]{\includegraphics[width = 0.45\textwidth]{plot/vis_PEMS03.pdf}} \hspace{5mm}
%     % \hfill
%     % \hspace{10mm} 
% 	\subfloat[PEMS08]{\includegraphics[width = 0.45\textwidth]{plot/vis_PEMS08.pdf}} 
% \caption{Traffic flow forecasting visualizations between \baby and baselines on real-world systems (a) CHIBike, (b) T-Drive, (c) PEMS03, (d) PEMS08. The vertical coordinate denotes the traffic flow values.}
% \label{fg_vis_real}
% \end{figure*}

% \end{figure}

\paragraph{Prediction of Incidence Proportion} 

% We evaluate the performance of prediction on incidence proportions on real-world traffic flow systems by MAPE in Table \ref{T_ip} in Appendix \ref{App_full_results}. We can find that our \baby significantly outperforms baseline methods in most settings, indicating that our \baby can serve as an effective forecaster for special case monitoring and warning in real-world applications.

% put into appendix
In this section, we evaluate the performance of prediction on incidence proportions. 
Incidence Proportion (IP) \cite{IncidentRate2010} measures the probability of a special event (\eg the infection of epidemic diseases) in a certain period. It is calculate by ${\rm IP}= \frac{D_e}{D_o}$, where $D_e$ denotes the occurrence amounts of a certain event; $D_o$ denotes the total amounts of all monitored subjects in a specified period.
We examine the error rates of prediction on IP on real-world traffic flow systems. Specifically, we calculate IP by assigning the traffic flow of each station object at a time point to $D_e$ and assign the traffic flow summation of all stations in the whole area at a time point to $D_o$. We calculate IP for both forecasted and ground-truth flows and measure the predicted IP by MAPE. The results are reported in Table \ref{T_ip}. We can find that our \baby outperforms baseline methods in most settings. Our method also beats STGODE and MTGODE in most cases, which are specially designed for traffic flow forecasting using temporal and spatial information. This indicates that our \baby can serve as an effective forecaster for special case monitoring and warning in real-world applications.

\begin{table}[!t]
    \centering
    \setlength{\tabcolsep}{0.2pt}
    % \tiny	
    % \small
    \scriptsize
    % \footnotesize	
    \caption{Averaged MAPE (in \%) scores of the predicted Incidence Proportion on traffic flows. The best scores are in \textbf{boldface} and second best scores are in \underline{underline}.}
    \renewcommand\arraystretch{1.05}

    \begin{tabular}{p{40pt}<{\centering}| 
    p{28pt}<{\centering}  
    p{28pt}<{\centering}  p{28pt}<{\centering} 
    p{28pt}<{\centering}  p{28pt}<{\centering} 
    p{28pt}<{\centering}  p{28pt}<{\centering} 
    p{28pt}<{\centering}  p{28pt}<{\centering} }
       
    \toprule
    
    Method &T-Drive &NYCTaxi &CHIBike &PEMS03 &PEMS04 &PEMS07 &PEMS08\\    

    \midrule
    LatentODE	&\textbf{13.9}	&188.8	&15.4	&20.5	&\underline{14.0}	&19.0	&12.0	\\
    GNS	&78.0	&\underline{44.7}	&\textbf{9.9}	&31.3	&26.2	&24.9	&19.7	\\
    NDCN	&120.1	&59.4	&\underline{14.6}	&35.6	&27.1	&22.4	&22.5	\\
    TREAT	&49.5	&177.9	&69.2	&42.9	&49.2	&20.7	&48.0	\\
    STGODE	&38.3	&35.6	&40.6	&17.6	&13.9	&11.0	&10.9	\\
    MTGODE	&253.9	&76.5	&102.7	&\underline{14.4}	&100.1	&\textbf{9.3}	&\underline{10.8}	\\
    \midrule
    \baby	&\underline{47.1}	&\textbf{30.1}	&17.6	&\textbf{14.1}	&\textbf{12.8}	&\underline{10.5}	&\textbf{8.8}	\\
    
    \bottomrule
    \end{tabular}
    \label{T_ip}
\end{table}

\section{Conclusion and Limitations}
\label{6}

% In this paper, we proposed a novel dynamics modeling framework consisting of two stages to learn generalizable dynamics. Firstly, we generate and collect benchmark datasets from simulated dynamics systems and realistic observations to pre-train a foundation model which can learn dynamics-enriched representations for trajectories. Secondly, we adopt the encoder (decoder) of the pre-trained foundation model as a representation learner to further learn the dynamics model. We model the hidden dynamics by solving the ODE initial value problem with a GNN-based neural ODE learner in the hidden space. In experiments, we conducted long/short term forecasting on in-domain and cross-domain settings. The empirical results demonstrate that our proposal beats 5 strong dynamics modeling methods and output generalizable dynamics. 

In this paper, we propose a generalized pre-trained dynamics encoder \baby which induce a latent space where the governing dynamics could be captured more easily than in the original observation space. During pre-training we collect 152 sets of dynamics observations from both synthetic and real-world systems as pre-training corpora. Specifically, we pre-train \baby by minimizing the M-Lyapunov exponent objective to constrain the chaotic degree of dynamics from the embedded observations to induce a latent space where the encoded observations can capture more stable and well-structured dynamics, thereby facilitating more effective and tractable latent dynamics modeling than in the original space. We also incorporate the reconstruction and forecasting objectives as auxiliary to prevent from learning an over-smoothed space and meanwhile to enhance the forecasting capacity for downstream dynamics modeling and forecasting tasks.
We also present the usage of fine-tuning \baby to learn specific dynamics. The initial states could be encoded by the pre-trained \baby and then used to learn dynamics in latent space by integrating using a GNN-based ODE learner. The integrated solutions are then decoded and adopted to fine-tune \baby by minimizing the forecasting loss against the ground-truth observation sequence. 
We conducted empirical studies on short/long-term forecasting under in-domain and cross-domain settings for validation. The results indicate the effectiveness and generalizability of our \baby in both synthetic and real-world systems, thereby \baby could be adopted as an effective representation learner for extracting governing dynamics when modeling in the original space is intractable.

Our method still has several limitations: the problems of catastrophic forgetting and pre-training corpus amount imbalance may affect the overall pre-training performance. Besides, how to integrate system-specific interacting graphs have not been fully explored.

\bibliography{PDEDER}
\bibliographystyle{IEEEtran}

\medskip

% \clearpage
% \appendix
\appendices
\label{7}

\section{Calculation of maximal Lyapunov Exponent}
\label{App_mle}

Given the embedded hidden states $\mathbf{z}_t$ of each time point $t\in[T]$ (we omit other subscripts for brevity), the corresponding M-Lyapunov exponent is calculated as below:
\begin{enumerate}
    \item We search the nearest neighbor $\mathbf{z}_{i^\prime}$ for each state point $\mathbf{z}_i$ using cosine similarity, where $i^\prime = \arg\min_{j} {\mathrm{cos}}<\mathbf{z}_i,\mathbf{z}_{j}>$. We denote $d_i(0)={\mathrm{cos}}<\mathbf{z}_i,\mathbf{z}_{i^\prime}>$ as the initial distance from $\mathbf{z}_i$ to its nearest neighbor $\mathbf{z}_{i^\prime}$ at time $t=0$.
    
    \item Following \cite{rosenstein1993practical}, we calculate the distance of each pair $(\mathbf{z}_i,\mathbf{z}_{i^\prime})$ after $t$ time points and denote it as $d_i(t)$.
    
    \item We formulate an averaged line $\mathbf{y}=\{y_t\}_{t=1}^{T_{max}}$ where $y_t=\frac{1}{\Delta t} <\ln d_i(t)>$, $<\cdot>$ denotes the averaged values of all $i$; $\Delta t$ denotes the sampling period of sequence. We use a differential least squares method $f_{ls}(\mathbf{y})$ to fit the line $\mathbf{y}$ and adopt the fitted slope as the approximated M-Lyapunov exponent.
    % \item Then we can use a differential least squares method to fit line $\mathbf{y}$ and the slope is the approximated maximal Lyapunov exponent. We calculate the slope $\mathbf{w}$ using the analytical solution: $\textbf{w}=(\mathbf{o^\top o})^{-1}\mathbf{o}^\top \mathbf{y}$, where $\mathbf{o}$ denotes the coordinate vector. Then we obtain the maximal Lyapunov Exponent by the slope solution $\lambda = \mathbf{w} = f_{ls}(\mathbf{y})$.
\end{enumerate}
 We refer to the readers for more details in \cite{rosenstein1993practical}.
 
By denoting the calculation of the averaged line $\mathbf{y}$ by $f_l(\cdot)$: $\mathbf{y} = f_{l}(\mathbf{z})$, we can calculate the gradients of the \baby encoder parameter $\mathbf{W}_e$ with respect to the MLE objective by:
\begin{equation}
    \frac{\partial \mathcal{L}_{mle}}{\partial \mathbf{W}_{e}} = \frac{\partial \mathcal{L}_{mle}}{\partial f_{ls}(\mathbf{y})} \cdot\frac{\partial f_{ls}(\mathbf{y}) }{\partial f_l(\mathbf{z})} \cdot \frac{\partial f_l(\mathbf{z})}{\partial \mathbf{W}_e}. \nonumber
\end{equation}

\section{Non-Invariance proof of maximal Lyapunov exponent under the encoding-decoding paradigm of \baby}
% coordinate transformation}
\label{App_proof}

As mentioned in \cite{MLEinvariance2001}, the Lyapunov exponents are invariant under invertible transformations of variables, and non-invertible transformations which can map the trajectories of two different systems to each other. While in our scenario, the above characteristics doesn't hold for our pre-trained encoder and decoder, therefore the M-Lyapunov exponent between the original observations and the embedded representations are not identical. We present detailed proof in what follows.

For brevity, we use $\mathbf{x}$ to denote the observations and denote the encoder and decoder of our pre-trained \baby by $f_e(\cdot)$ and $f_d(\cdot)$. 

When all the hidden states $\{\mathbf{x}_t\}_{t=1}^T$ of the observation sequence are available, it is obvious that our pre-trained encoder and decoder is not invertible and cannot transform the observed sequence to each other even with a reconstruction objective. The objectives can force the model to maintain original (reconstruction) and future (forecasting objective) sequence patterns, but it cannot strictly induce a mutually inverse encoder and decoder. This is caused by the natural characteristics of some architectures such as multi-head attention mechanism in the transformer as mentioned in \cite{sukthanker2022generative}. The non-bijectivity and information aggregation make it impossible to recover the input from the output, even with ideal training. In short, the encoder and decoder are not mutually inverse and the $\mathbf{x}=\hat{\mathbf{x}}$ cannot hold in our framework.

When only the initial state $\mathbf{x}_0$ are available, the sequence states in the original space $\{\mathbf{x}_1, \ldots,\mathbf{x}_T\}$ and states in the latent space  $\{\mathbf{z}_1, \ldots,\mathbf{z}_T\}$ could be calculated by solving the initial value problems:
\begin{equation}
% \begin{split}    
    \mathbf{x}_i = \mathbf{x}_0 + \int_0^i h(\mathbf{x}_t)dt,\ \ \ \ 
    \mathbf{z}_i = \mathbf{z}_0 + \int_0^i g(\mathbf{z}_t)dt, \nonumber
% \end{split}    
\end{equation}
where $\mathbf{z}_0=f_e(\mathbf{x}_0)$; $h(\cdot)$ and  $g(\cdot)$ denotes the corresponding governing equations. The reconstructed $\hat{\mathbf{x}}$ is decoded by $\hat{\mathbf{x}}=f_d(\mathbf{z})$.

According to \cite{MLEinvariance2001}, if the integrated states $\mathbf{x}$ learnt from $\mathbf{x}_0$ in the original space are identical to $\hat{\mathbf{x}}$ obtained by decoding from latent variable $\mathbf{z}$, the Lyapunov exponents computed on $\mathbf{x}$ and $\mathbf{z}$ remain invariant under the transformation. The invariance holds when  $\mathbf{x}_i=\hat{\mathbf{x}}_i, \forall i \in [T]$. While this constraint cannot hold under our framework, the detailed proof are listed below:

Assume that the hidden dynamics could be described by a simple linear equation:
\begin{equation}
    \frac{d\mathbf{x}_t}{dt} = h(\mathbf{x}_t)=\mathbf{W}_1^\top\mathbf{x}_t, \ \ \ 
    \frac{d\mathbf{z}_t}{dt} = g(\mathbf{z}_t)=\mathbf{W}_2^\top\mathbf{z}_t. \nonumber
\end{equation}
Then $\mathbf{x}_i$ and $\mathbf{z}_i$ could be calculated by integration:
$\mathbf{x}_i = \mathbf{x}_0 + \int_0^i h(\mathbf{x}_t)dt = \mathbf{x}_0\cdot e^{\mathbf{W_1^\top}i}$ and $
\mathbf{z}_i = \mathbf{z}_0 + \int_0^i g(\mathbf{z}_t)dt = \mathbf{z}_0\cdot e^{\mathbf{W_2^\top}i} $.
% \begin{equation}
% \begin{split}    
%     \mathbf{x}_i &= \mathbf{x}_0 + \int_0^i h(\mathbf{x}_t)dt = \mathbf{x}_0\cdot e^{\mathbf{W_1^\top}i} \\
%     \mathbf{z}_i &= \mathbf{z}_0 + \int_0^i g(\mathbf{z}_t)dt = \mathbf{z}_0\cdot e^{\mathbf{W_2^\top}i} \nonumber
% \end{split}
% \end{equation}
Then determining whether $\mathbf{x}_i=\hat{\mathbf{x}}_i$ holds can be transformed to determining whether $\mathbf{x}_0\cdot e^{\mathbf{W}_1^\top i} = f_d(\mathbf{z}_0\cdot e^{\mathbf{W}_2 i })$ holds.
\begin{enumerate}
    \item When $i=0$, the above equation is transformed to $\mathbf{x}_0=f_d(f_e(\mathbf{z_0}))$. This means that the encoder and decoder are invertible and could reconstruct the input. While, the encoder and decoder of transformer-based language models are invertible and could not reconstruct the input injectively. As mentioned in \cite{sukthanker2022generative}, the information aggregation and non-injectivity lead to non-invertibility of attention mechanism in transformer. And \cite{zhu2023making, Kitaev2020Reformer} also developed reversible transformers, indicating the non-invertibility of transformers. Therefore, $\mathbf{x}_0=f_d(f_e(\mathbf{z_0}))$ does not hold when $i=0$.

    \item When $i \neq 0$, we first assume $\mathbf{x}_0\cdot e^{\mathbf{W}_1^\top i} = f_d(\mathbf{z}_0\cdot e^{\mathbf{W}_2^\top i })$ holds. Due to the non-injectivity, there may exist $\mathbf{x}_0^\prime \neq \mathbf{x}_0$ that satisfies $f_e(\mathbf{x_0}) = f_e(\mathbf{x}_0^\prime)$. Therefore, $f_d(f_e(\mathbf{x}_0\cdot e^{\mathbf{W}_2^\top i})) = f_d(f_e(\mathbf{x}_0^\prime\cdot e^{\mathbf{W}_2^\top i}))$. Then we have $\mathbf{x}_0\cdot e^{\mathbf{W}_1^\top i} = \mathbf{x}_0^\prime\cdot e^{\mathbf{W}_1^\top i}$. While we have known that $\mathbf{x}_0 = \mathbf{x}_0^\prime$, therefore, the assumption $\mathbf{x}_0\cdot e^{\mathbf{W}_1^\top i} = f_d(\mathbf{z}_0\cdot e^{\mathbf{W}_2^\top i })$ does not hold. 
\end{enumerate}
According to the above proof, the Lyapunov exponent invariance condition does not hold under our framework, indicating that the Lyapunov exponents calculated by the original states $\mathbf{x}$ and the latent representations $\mathbf{z}$ are not invariant under the transformation.
$\square$

\vfill

\end{document}